\documentclass[pdflatex,sn-mathphys-num]{sn-jnl}

\usepackage{amsmath,amssymb,amsthm,mathtools,amsfonts}
\usepackage{multirow}
\usepackage{mathrsfs}
\usepackage[title]{appendix}
\usepackage{xcolor}
\usepackage{textcomp}
\usepackage{manyfoot}
\usepackage{lineno}
\usepackage{listings}
\usepackage{enumitem}
\usepackage[mathscr]{eucal}
\usepackage{tabularx}
\usepackage{array}
\usepackage{float}
\usepackage{siunitx}
\usepackage{changepage}
\usepackage{soul}
\usepackage{graphicx}
\usepackage{booktabs}

\usepackage{hyperref}
\usepackage{cleveref}

\hypersetup{
  colorlinks=true,
  linkcolor=blue!55!black,
  citecolor=blue!55!black,
  urlcolor=blue!55!black
}

\newcolumntype{Y}{>{\centering\arraybackslash}X}
\newcolumntype{L}{>{\raggedright\arraybackslash}X}
\newcolumntype{C}{>{\centering\arraybackslash}X}
\newcolumntype{R}{>{\raggedleft\arraybackslash}X}

\theoremstyle{definition}

\numberwithin{equation}{section}

\begin{document}

\title{FTU-Seek: Foundation Model-Guided Hard-Negative Learning for Sparse Functional Tissue Unit Segmentation}

\author[1,2]{\fnm{Zonghao} \sur{Liu}}\email{liuzonghao42@gmail.com}
\equalcont{These authors contributed equally to this work.}

\author[3,4]{\fnm{Lei} \sur{Su}}\email{sulei2023@ia.ac.cn}
\equalcont{These authors contributed equally to this work.}

\author[5]{\fnm{Jiguang} \sur{Yu}}\email{jyu678@bu.edu}
\equalcont{These authors contributed equally to this work.}

\author[3,4]{\fnm{Xuqing} \sur{Geng}}\email{gengxuqing2024@ia.ac.cn}

\author*[6]{\fnm{Louis Shuo} \sur{Wang}}\email{wang.s41@northeastern.edu}

\author[2,7]{\fnm{Jianmin} \sur{Wang}}\email{wangjm8605@163.com}

\author*[1,7,8]{\fnm{Jingfeng} \sur{Liu}}\email{drjingfeng@126.com}

\affil[1]{%
  \orgdiv{Department of Hepatopancreatobiliary Surgery},
  \orgname{Clinical Oncology School, Fujian Medical University},
  \orgaddress{\city{Fuzhou}, \postcode{350014}, \country{China}}
}

\affil[2]{%
  \orgdiv{Innovation Center for Cancer Research},
  \orgname{Clinical Oncology School, Fujian Medical University},
  \orgaddress{\city{Fuzhou}, \postcode{350014}, \country{China}}
}

\affil[3]{%
  \orgdiv{Institute of Automation},
  \orgname{Chinese Academy of Sciences},
  \orgaddress{\city{Beijing}, \postcode{100190}, \country{China}}
}

\affil[4]{%
  \orgname{University of Chinese Academy of Sciences},
  \orgaddress{\city{Beijing}, \postcode{100190}, \country{China}}
}

\affil[5]{%
  \orgdiv{College of Engineering},
  \orgname{Boston University},
  \orgaddress{\city{Boston}, \state{MA}, \postcode{02215}, \country{USA}}
}

\affil[6]{%
  \orgdiv{Department of Mathematics},
  \orgname{Northeastern University},
  \orgaddress{\city{Boston}, \state{MA}, \postcode{02115}, \country{USA}}
}

\affil[7]{%
  \orgdiv{Fujian Key Laboratory of Advanced Technology for Cancer Screening and Early Diagnosis},
  \orgname{Fujian Cancer Hospital},
  \orgaddress{\city{Fuzhou}, \postcode{350014}, \country{China}}
}

\affil[8]{%
  \orgdiv{Key Laboratory of Cancer Metabolism},
  \orgname{National Health Commission of the People's Republic of China},
  \orgaddress{\city{Fuzhou}, \postcode{350014}, \country{China}}
}

\abstract{
\textbf{Background/Objectives:}
Functional tissue units (FTUs), including tertiary lymphoid structures (TLSs), blood vessels, and glands, encode localized immune, vascular, and epithelial organization in histopathology. Accurate quantification of these structures is important for studying tissue architecture and disease-associated tissue organization. However, FTUs are frequently sparse, heterogeneous, and surrounded by large amounts of morphologically similar background tissue, making automated segmentation in whole-slide images (WSIs) challenging. We therefore developed FTU-Seek, a pathology foundation model-guided framework that treats morphology-aware negative-patch selection as a key component of sparse FTU segmentation.
\textbf{Methods:}
FTU-Seek uses frozen multi-depth features from the UNI pathology foundation model to train a patch-level classifier that distinguishes FTU-containing from FTU-absent tissue. Target-absent patches are subsequently ranked according to their predicted target-containing probabilities, and the highest-scoring hard negatives are selected through a static Top$K$ strategy to construct compact segmentation training sets. 
The framework was evaluated using five-fold cross-validation and internal test cohorts across TLS, blood-vessel, and gland segmentation tasks, with an additional independent 30-WSI held-out cohort for TLS.
Positive-only, all-tissue, random-negative, and matched random Top$K$ sampling strategies served as comparators. Segmentation-derived phenotypes were further explored in external TCGA cohorts.
\textbf{Results:}
The patch-level classifiers achieved mean validation AUCs of 95.92\%, 90.11\%, and 98.16\% for TLS, blood vessel, and gland classification, respectively.
For TLS segmentation, the pre-specified Top1000 configuration retained 27.6\% of the all-tissue training workload and achieved a slide-level Dice of 76.69 ± 11.89\% on the independent 30-WSI held-out cohort. Compared with matched random Top1000 sampling, it improved Dice by 3.72 percentage points (95\% CI, 2.05–5.38).
Blood vessel and gland segmentation achieved performance approaching all-tissue training while reducing the retained training workload by approximately one-half and one-third, respectively. Compared with matched random sampling, classifier-guided hard-negative selection produced the greatest improvements for sparse and morphologically ambiguous FTUs. Exploratory TCGA analyses further showed associations of TLS phenotypes with overall survival, vascular phenotypes with overall survival and microvascular invasion, and glandular phenotypes with clinicopathological characteristics.
\textbf{Conclusions:}
FTU-Seek demonstrates that pathology foundation models can support sparse FTU segmentation not only through feature representation but also through morphology-aware construction of segmentation training sets. By prioritizing informative hard negatives, the framework reduces redundant segmentation-training workload while maintaining competitive segmentation performance and supporting quantitative tissue phenotyping from routine histopathology.
}



\maketitle

\section{Introduction}

Whole-slide images (WSIs) are increasingly used not only for diagnosis, grading, and~treatment planning, but~also for quantitative analysis of tissue architecture and the tumor microenvironment~\cite{kumar2020whole, bera2019artificial,bian2021immunoaizer}. Beyond~large histological compartments, WSIs contain spatially localized microscopic structures that reflect complementary biological processes. Glandular structures encode epithelial differentiation and architectural organization~\cite{bulten2020automated, strom2020artificial,wang2025analysis}, blood vessels reflect angiogenesis and vascular remodeling~\cite{lugano2020tumor, seraphin2025artificial}, and~tertiary lymphoid structures (TLSs) represent organized antitumor immune activity~\cite{sautes2019tertiary, cabrita2020tertiary,wang2025analysis1, michot2026beyond}. In~this study, we use the term functional tissue units
 (FTUs) to describe these histologically identifiable multicellular structures whose morphology, abundance, and~spatial organization are linked to epithelial, vascular, or~immune functions. Accurate FTU segmentation may therefore provide a practical route for extracting interpretable image-derived phenotypes from routine histopathology slides~\cite{diao2021human,liang2025global,su2026gcunet, ye2026foundation}.

\begin{figure}[H]
    \includegraphics[width=\columnwidth]{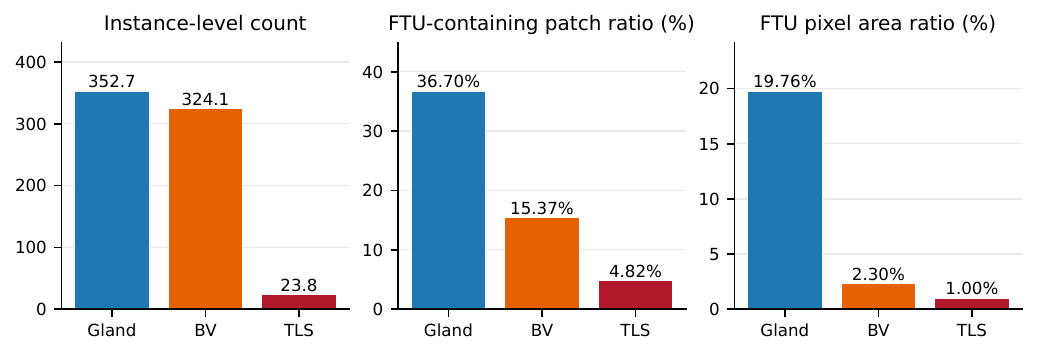}
    \caption{Quantitative assessment of spatial sparsity across representative functional tissue units (FTUs). Glands, blood vessels (BVs), and~tertiary lymphoid structures (TLSs) were compared at the instance, patch, and~pixel levels. Instance-level sparsity was assessed using FTU instance counts; the proportion of FTU-containing tissue patches measured patch-level sparsity; and pixel-level sparsity was quantified using the FTU pixel area ratio at a spatial resolution of 1 \unit{\micro\metre}/pixel. This figure reports statistics calculated from the complete cohort by pooling the development and independent test~sets.}
    \label{fig:sparse_quantification}
\end{figure}

\begin{figure}[H]
     \includegraphics[width=\columnwidth]{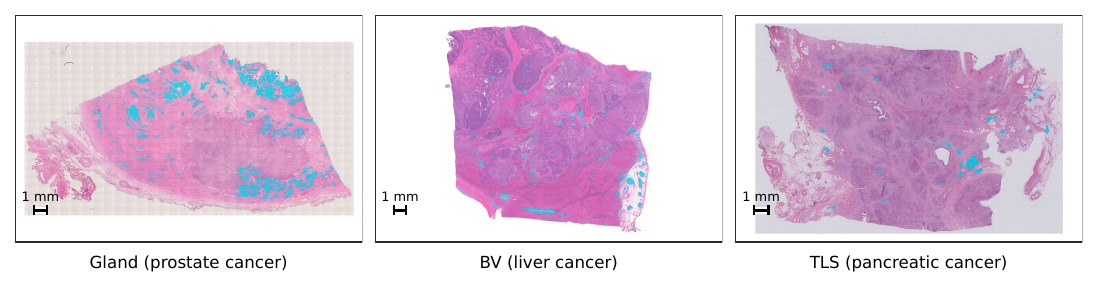}
    \caption{Representative
 spatial distributions of annotated functional tissue unit (FTU) regions in whole-slide images. Annotated FTU regions are shown in cyan for glandular structures in prostate cancer, blood vessels in liver cancer, and~tertiary lymphoid structures in pancreatic~cancer.}
    \label{fig:sparse_visualization}
\end{figure}

A central challenge in FTU analysis is that these structures are frequently sparse, heterogeneous, and~embedded within large amounts of target-absent tissue. As~shown in
 \Cref{fig:sparse_quantification}, the~representative FTUs studied here differ substantially at the instance, patch, and~pixel levels. TLSs and blood vessels occupy only a small fraction of tissue patches and total tissue area~\cite{schumacher2022tertiary,yu2026pattern,xia2025learnable, hamidinekoo2021automated}, whereas glandular structures are more abundant but still exhibit considerable architectural and morphological variability~\cite{sirinukunwattana2017gland,yu2026size,chen2017dcan}. Their spatial distributions are also highly uneven across prostate cancer, liver cancer, and~pancreatic cancer WSIs, as~illustrated in \Cref{fig:sparse_visualization}. Manual delineation by pathologists is possible but labor-intensive, particularly when FTUs are small, fragmented, boundary-ambiguous, or~visually confounded by surrounding tissue~\cite{niazi2019digital, sevim2026s,wang2026algebraic,al2026pathology, xia2026beyond}. These characteristics make FTU segmentation a sparse-structure learning problem rather than a conventional segmentation task involving large and spatially continuous~regions.

Deep learning-based WSI segmentation has achieved strong performance for relatively large and spatially continuous compartments, such as tumor parenchyma, necrosis, and~stroma~\cite{hosseini2024computational, al2023applications, kather2019predicting, saltz2018spatial,wang2026damage,campanella2019clinical}. Sparse FTU segmentation imposes a fundamentally different constraint. Target-containing patches often represent only a small minority of all tissue patches, whereas the substantially larger target-absent pool contains both redundant easy negatives and a limited subset of morphologically target-like regions that can trigger false-positive (FP) predictions~\cite{van2021hooknet,lee2024ensemble}. Training on all tissue patches is therefore inefficient because a large proportion of the negative samples contributes little additional information. Random negative sampling reduces the training workload, but~may fail to retain the rare background regions that most closely resemble the target and are therefore most important for FP suppression. Accordingly, sparse FTU segmentation is not only a pixel-level prediction problem, but~also a WSI-scale training-set construction problem: the model must be exposed to sufficiently informative negative tissue without being dominated by redundant background~patches.

Existing imbalance-handling strategies only partially address this challenge. Loss-based methods, including class-weighted objectives, Dice-based losses, and~focal loss, increase the optimization contribution of under-represented or difficult pixels~\cite{milletari2016v, lin2017focal,wang2026breakdown}. Sampling-based approaches rebalance the training distribution through oversampling, undersampling, or~hard-example selection. Online hard example mining prioritizes high-loss samples produced by the current model and has been influential in computer vision~\cite{shrivastava2016training}. In~computational pathology, hard-negative mining has also been applied to histopathology classification and weakly supervised WSI analysis~\cite{li2019deep,yu2026beyond,huang2024hard}. However, many previous methods have primarily been developed for image- or slide-level classification, or~for dense prediction settings in which candidate samples can be evaluated within a manageable field of view. For~gigapixel WSI segmentation, repeatedly screening the complete negative pool during pixel-level training can be computationally expensive. 
{A remaining practical challenge is to identify, before~segmentation training, a~compact and reproducible subset of target-absent patches enriched for morphologically confusing hard negatives.}
{This setting is distinct from active learning, which typically selects
unlabeled samples for additional expert annotation and iteratively expands
the labeled training set, and~from online hard-example mining, which
dynamically identifies high-loss or misclassified samples during model
optimization~\cite{shrivastava2016training}. In~contrast, FTU-Seek
operates on an already annotated WSI patch pool. It ranks FTU-absent
patches before segmentation training and constructs a compact, fixed, and~reproducible training set without requesting additional annotations.}

Pathology foundation models (PFMs) {provide a promising basis for this selection problem. Models such as
CTransPath, UNI, Virchow, and~Prov-GigaPath have been used to learn
transferable histomorphological representations, while vision--language
models such as CONCH incorporate semantic supervision from paired
biomedical text
~\cite{wang2022transformer,yu2026rigorous,chen2024towards,
vorontsov2024foundation,xu2024whole,lu2024visual,yu2026microscopic}.
These representations have demonstrated broad utility across cancer types
and downstream prediction tasks~\cite{campanella2025clinical}. Related
studies have also explored pathology patch selection, representative
region selection, and~hard-example mining
~\cite{li2019deep,yu2026beyond,huang2024hard}. Accordingly, PFMs and
patch-selection strategies have been investigated in related settings.} 

{Here, we propose FTU-Seek, 
 a~PFM-guided hard-negative learning
framework for efficient segmentation of sparse FTUs in WSIs. FTU-Seek
uses frozen multi-depth PFM features to train a patch-level classifier,
which ranks an already annotated FTU-absent patch pool according to
target-like morphology. A~static per-WSI Top$K$ rule is then used to
select the highest-scoring negative patches and construct a compact,
workload-controlled segmentation training set. Thus, FTU-Seek integrates
PFM-based morphological representation, target-specific hard-negative
ranking, and~static training-set construction for sparse FTU segmentation.}

The main contributions of this study are summarized as~follows:
\begin{itemize}
\item {We develop a PFM-guided training-set construction strategy that
integrates frozen multi-depth representations, target-specific ranking of
annotated FTU-absent patches, and~static per-WSI Top$K$ selection to reduce
redundant background sampling in sparse FTU segmentation.}
\item We systematically evaluate classifier-guided hard-negative selection across TLS, blood-vessel, and~gland segmentation tasks with distinct sparsity and morphological profiles, using positive-only, all-tissue, random-negative, and~matched random Top$K$ strategies to assess the trade-off between segmentation performance and retained \mbox{training~workload.}

\item We further demonstrate the downstream utility of FTU-Seek through exploratory applications to external TCGA cohorts, showing that segmentation outputs can be transformed into quantitative phenotypes describing immune, vascular, and~glandular tissue organization.
\end{itemize}

Because the selected negative pool is fixed before segmentation training, this
construction reduces the number of patches that must be stored, loaded, and~processed during the segmentation stage. The~resulting benefit is therefore
an expected reduction in training-data workload and data-transfer burden,
rather than a directly measured reduction in wall-clock time, GPU-memory
usage, or~energy~consumption.

\section{Materials and~Methods} 

\subsection{Study Design and~Datasets}
To develop FTU-Seek for precise spatial segmentation, our methodology was primarily driven by highly granular, instance-level morphological analysis. We retrospectively constructed three task-specific H\&E-stained WSI cohorts from Fujian Cancer Hospital (FCH). For~each task, we purposely curated a representative subset of WSIs designed to encompass a wide spectrum of tissue architectures and staining profiles, thereby maximizing morphological heterogeneity. Through exhaustive, pixel-level manual delineation of H\&E-stained whole-slide images, we extracted three datasets of FTUs, establishing annotations used to dichotomize all derived image patches into FTU-containing and FTU-absent categories. Specifically, the~TLS dataset comprised 2079 FTU-containing patches and 41,094 FTU-absent patches from patients with pancreatic ductal adenocarcinoma (PAAD); the gland dataset included 11,670 FTU-containing patches and 20,128 FTU-absent patches from patients with prostate adenocarcinoma (PRAD); and the blood vessel dataset included 4005 FTU-containing patches and 22,048 FTU-absent patches from patients with intrahepatic cholangiocarcinoma. Each patient contributed one WSI. For~each task, WSIs were randomized into a model development set and an independent test set at a 4:1 ratio before patch extraction. 
The resulting cohorts contained 10 patients and 10 TLS WSIs (8 for development and 2 for internal testing), 9 patients and 9 blood-vessel WSIs (7 for development and 2 for internal testing), and~10 patients and 10 gland WSIs (8 for development and 2 for internal testing). Data partitioning was performed at the patient/WSI level and was independent of patch-level sampling. Patch extraction and subsequent positive/negative patch selection were conducted within the predefined development or test cohort, without~using patch-level information to determine the split. Given the limited number of internal-test WSIs, these results were interpreted cautiously and were complemented by the independent 30-WSI held-out TLS~evaluation.

The development sets—which provided a morphologically dense corpus of training patches—underwent five-fold cross-validation to optimize model hyperparameters, while the independent test sets were completely held out for final segmentation performance evaluation. To~further assess the robustness of the segmentation strategies beyond the original two-slide internal test set, we constructed an additional held-out cohort of 30 TLS whole-slide images from the same institution. These WSIs were not used for model training, model selection, hyperparameter tuning, Top$K$ selection, or~threshold optimization. The~30-slide cohort was evaluated only after the segmentation configurations had been pre-specified based on the development-set experiments. For~each method, predictions from the five cross-validation models were combined as an ensemble, and~a fixed segmentation threshold of 0.5 was used for all~evaluations.

To investigate the downstream utility of FTU segmentation, diagnostic WSIs and matched clinicopathological data from public TCGA cohorts were downloaded from the NIH Genomic Data Commons Data Portal (\url{https://portal.gdc.cancer.gov}) \cite{weinstein2013cancer}. 
One diagnostic WSI was included per patient.
Three task-specific external analyses were performed. 
For TLS phenotyping, TCGA-READ, TCGA-ESCA, and~TCGA-STAD included 146, 148, and~326 patients, with~26, 62, and~132 deaths, respectively.
TLS phenotypes were quantified in TCGA-READ, TCGA-ESCA, and~TCGA-STAD to assess immune-structure abundance, size composition, and~spatial organization. 
For vascular phenotyping, TCGA-LIHC included 337 patients with 121 deaths for overall survival (OS) analysis and the microvascular invasion (MVI) analysis.
Blood-vessel phenotypes were quantified in TCGA-LIHC.
For glandular phenotyping, TCGA-PRAD included 401 patients, with~395, 372, and \mbox{342 patients} having available pathological T stage, surgical margin, and~biochemical recurrence (BCR) information, respectively.
Glandular phenotypes were quantified and evaluated in relation to Gleason grade group and the available clinicopathological endpoints described above. TCGA cases were retained when a diagnostic WSI, matched clinical endpoint, and~valid tissue-level segmentation output were available. Additionally, we excluded patients with concurrent other malignancies, those who underwent resection for tumor recurrence, and~those with a follow-up duration of less than one month.
These analyses were designed as exploratory downstream applications of FTU-Seek-derived segmentation outputs rather than as external validation of segmentation performance or definitive clinical prediction~analyses.

All procedures and protocols in this study adhered to the principles of the Declaration of Helsinki. Ethical approval for the retrospective FCH cohorts was obtained from the Ethics Committee of Fujian Cancer Hospital on 14 August 2024 (No. SQ2026-050). The~requirement for informed consent was waived because the study used fully de-identified retrospective clinical data. Use of public datasets from TCGA was deemed exempt from ethics committee review because these data are de-identified and publicly~available.

\subsection{WSI Annotation and~Preprocessing}
All WSIs were acquired at $20\times$ or $40\times$ magnification. For~pathological annotation, the~boundaries of the target FTUs---namely TLSs, glands, and~blood vessels---were manually delineated by three trained annotators using Aperio ImageScope v12.4.6 (Leica Biosystems, Wetzlar, Germany), with~one annotator assigned to each FTU type. All annotations were subsequently reviewed and finalized by a senior pathologist (Y.K.) to ensure consistency with the predefined histopathological criteria described below. As~annotations were not independently duplicated, inter-observer agreement was not assessed. Annotators were blinded to clinical outcome information during~annotation.

\begin{itemize}
\item TLSs.
 TLSs were defined as dense aggregations of lymphocytes within non-lymphoid tissues, while loosely distributed inflammatory infiltrates lacking clear borders or cohesive architecture were excluded~\cite{amisaki202533,cai2026optimal,ling2022prognostic}.

\item Blood vessels. The annotation of vascular structures depended on the presence of an endothelial lining. Vascular contours were delineated along the outer wall, encompassing the endothelial lining and any associated structural layers (such as smooth muscle). To~ensure a comprehensive vascular assessment, the~annotations incorporated a diverse range of vessel types, spanning from arterioles with distinct smooth muscle layers to portal and hepatic veins, which frequently lack conspicuous smooth muscle tissue. To~prevent the model from learning artifacts, tissue tearing spaces mimicking vascular lumens but lacking an endothelial lining were explicitly excluded~\cite{kumar2014robbins,yu2026age,o2023wheater}.

\item Glands. The boundaries of prostatic glands were precisely traced at the epithelial-stromal junction, excluding the surrounding fibromuscular stroma. These annotations encompassed both benign glands---marked by an intact bilayered architecture (inner secretory and outer basal cells) alongside intraluminal corpora amylacea---and malignant acini, characterized by basal cell loss, crowded or cribriform growth patterns, and~enlarged hyperchromatic nuclei~\cite{epstein20162014,wang2026elliptic,dasgupta2022re}.
\end{itemize}

After annotation, foreground tissue regions were separated from the surrounding slide background using Otsu thresholding across all datasets~\cite{lu2021data}. To~balance morphological detail with computational efficiency, tissue regions were processed at a spatial resolution of 1 \unit{\micro\metre}/pixel, corresponding to approximately $10\times$ magnification~\cite{van2021hooknet,yu2026fullcovariance}. Each foreground tissue region was subsequently divided into non-overlapping image patches of $256 \times 256$ pixels. Before~model input, these patches were resized to $224 \times 224$ pixels. 

The same non-overlapping patch coordinates were retained throughout patch-level
classification, segmentation training, validation, and~testing. No additional
random cropping or overlapping patch generation was performed at later stages.
The image transformations applied during model input consisted only of resizing
the extracted patches to the model input size, conversion to tensors, and~channel-wise normalization using ImageNet mean and standard-deviation values.
No random flipping, rotation, color jittering, or~other data augmentation was
used for either the patch classifier or the segmentation~model.

Patches located entirely in slide background regions were excluded from subsequent analysis. For~patch-level classification, any patch overlapping the annotated target region was labeled as target-containing, whereas patches without target overlap were labeled as target-absent. This inclusive overlap criterion was selected because the auxiliary classifier was intended to identify all patches containing potential target morphology, including partial or boundary-crossing targets, for~subsequent hard-negative ranking. Using a minimum-overlap threshold could discard small or peripheral target regions and increase the risk of assigning target-containing context to the negative pool. These patch-level labels were used only for classifier training and negative-patch ranking, while segmentation training and evaluation were performed using the original pixel-level annotation masks. Thus, non-target pixels within a target-containing patch remained background during segmentation learning. The~same ImageNet mean and standard-deviation normalization was used for all classifier and segmentation inputs to reduce differences in input scale across~WSIs.

\subsection{Model~Development}

\begin{figure}[H]
    
    \includegraphics[width=\textwidth]{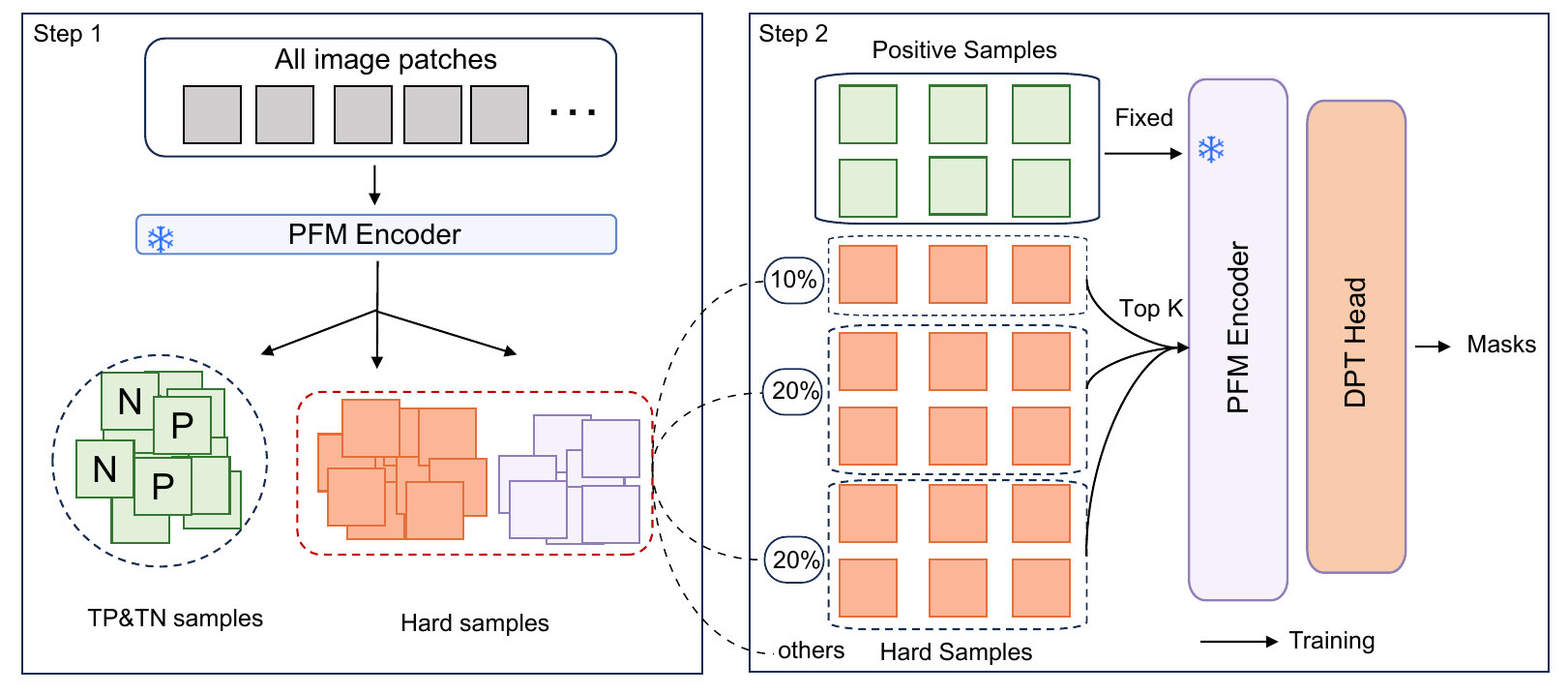}
    \caption{Overview of the proposed FTU-Seek framework. A~patch-level classifier first ranks FTU-absent tissue patches according to their predicted FTU-containing probabilities, and~classifier-derived TopK hard negatives are selected to construct compact segmentation training sets. The~framework is evaluated by comparing positive-only training, all-tissue training, random TopK sampling, and~classifier-guided FP-find TopK hard-negative~sampling.}
    \label{fig:framework}
\end{figure}

As shown in \Cref{fig:framework}, FTU-Seek was designed as a two-stage hard-negative learning framework for sparse FTU segmentation in WSIs. The~framework first uses a patch-level classifier to identify FTU-absent tissue patches that are likely to be confused with FTU-containing~patches. 

The pathology foundation model used in FTU-Seek was UNI v1. Its encoder was
kept frozen, and~multi-depth UNI v1 features were used as inputs to a
task-specific patch-level binary classifier. The~classifier was trained to
distinguish FTU-containing patches from FTU-absent patches using the
development-set training WSIs within each cross-validation fold. After~training, the~classifier assigned each FTU-absent patch a probability of being
FTU-containing. Within~each WSI, FTU-absent patches were ranked according to
this probability, and~the highest-scoring Top$K$ patches were selected as
hard negatives. These selected hard negatives were combined with all
FTU-containing patches to form the segmentation training set. The~resulting
training set was then used to optimize the segmentation decoder and prediction
head, while the UNI v1 encoder remained~frozen.

These classifier-derived hard negatives are then used to construct static TopK segmentation training sets. The~static strategy evaluates how the number of classifier-derived hard negatives affects segmentation performance and directly compares FP-find TopK sampling with positive-only training, all-tissue training, and~random TopK sampling under matched negative-sampling~budgets.

\subsubsection{Patch-Level Training Set Definition}

Stage I establishes a classifier-guided hard negative pool through patch-level FP screening. This stage consists of patch-level training set definition, classification network construction, and~classifier-derived hard negative ranking.
Let $\mathcal{P}$ denote the complete set of tissue patches extracted from the training WSIs. According to their overlap with the corresponding annotation masks, $\mathcal{P}$ is partitioned into a target-containing subset $\mathcal{P}^{+}$ and a target-absent subset
 $\mathcal{P}^{-}$:
\[
\mathcal{P}
=
\mathcal{P}^{+}
\cup
\mathcal{P}^{-},
\qquad
\mathcal{P}^{+}
\cap
\mathcal{P}^{-}
=
\varnothing.
\]
A patch
 is assigned to $\mathcal{P}^{+}$ if it overlaps with at least one annotated target region; otherwise, it is assigned to $\mathcal{P}^{-}$. These patch-level labels are used to train the classifier and to identify high-risk background patches for hard negative~initialization.

\subsubsection{Classification Network}

\begin{figure}[H]
    \centering
    \includegraphics[width=1.0\textwidth]{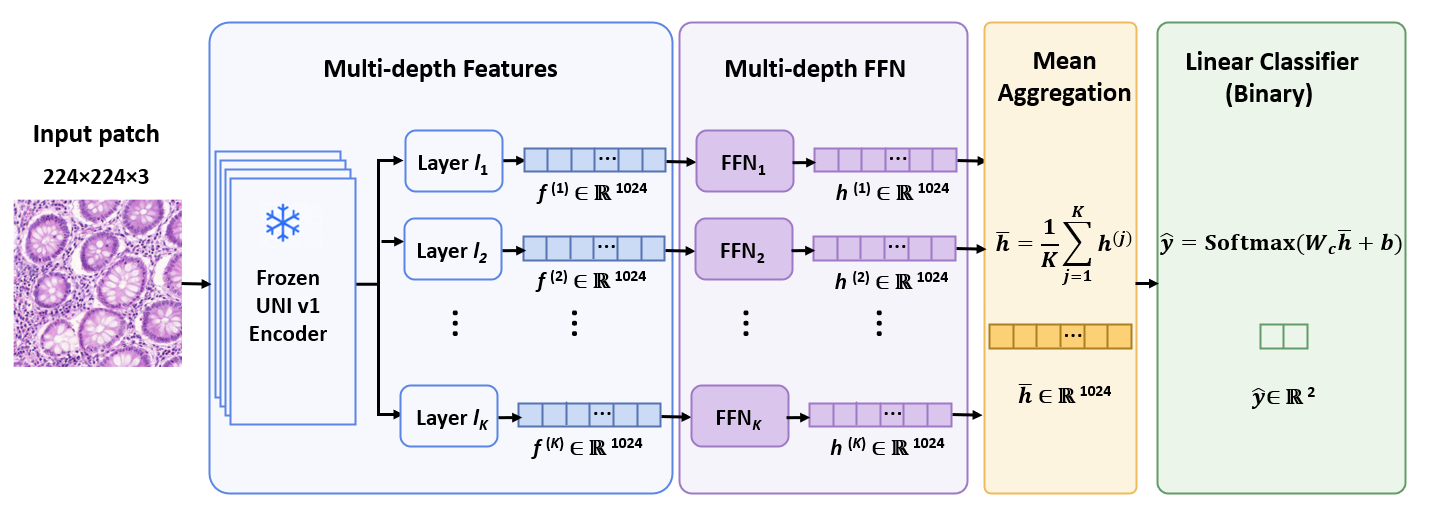}
    \caption{Architecture of the patch-level classification network. Multi-dimensional feature embeddings are extracted from different relative depths of the frozen UNI v1 encoder. Each feature is processed by a depth-specific feed-forward transformation head, and~the transformed representations are aggregated before binary patch-level~classification.}
    \label{fig:cls_network}
\end{figure}

The architecture of the patch-level classification network is illustrated in Figure~\ref{fig:cls_network}. For~each input patch, feature embeddings are extracted from $K$ selected relative depths of the frozen UNI v1 encoder~\cite{chen2024towards,wang2025multi}. Each depth-specific feature is independently processed by a dedicated feed-forward transformation head. The~transformed features are then aggregated into a fused patch representation, which is finally mapped to a binary prediction indicating whether the patch is target-containing or~target-absent.

Specifically, for~an input patch $x_i$, the~frozen UNI v1 encoder extracts $K$ feature embeddings, denoted as
$\mathbf{f}_i^{(j)}=E_{\mathrm{UNI}}^{(r_j)}(x_i)$ for $j=1,2,\ldots,K$. Here, $E_{\mathrm{UNI}}^{(r_j)}(\cdot)$ denotes the encoder output at the selected relative depth $r_j$, and~$\mathbf{f}_i^{(j)} \in \mathbb{R}^{1024}$.

Each depth-specific feature is independently transformed by a feed-forward head:
\[
\mathbf{h}_i^{(j)}
=
F_j\!\left(\mathbf{f}_i^{(j)}\right),
\qquad
j=1,2,\ldots,K,
\]
where $F_j(\cdot)$ denotes the feed-forward head for the $j$-th selected encoder depth. In~this study, $K=10$, with~relative encoder depths uniformly selected as $0.1,0.2,\ldots,1.0$. In~our implementation, each head consists of two fully connected layers with a ReLU activation, expanding the feature dimension from $1024$ to $4096$ and projecting it back to $1024$.

The transformed representations from all selected depths are fused by mean aggregation and mapped to binary class probabilities:
\begin{equation}
\hat{\mathbf{y}}_i
=
\operatorname{Softmax}
\left(
C
\left(
\frac{1}{K}
\sum_{j=1}^{K}
\mathbf{h}_i^{(j)}
\right)
\right),
\qquad
\hat{\mathbf{y}}_i \in \mathbb{R}^{2},
\label{eq:cls_prediction}
\end{equation}
where $C(\cdot)$ denotes the final linear classification layer. The~two output dimensions correspond to the target-absent and target-containing classes, respectively.
After training, the~patch-level classifier is applied to the training patch set. For~each patch $x_i$, the~predicted patch label is defined as
\[
\hat{c}_i
=
\arg\max_{c \in \{0,1\}}
\hat{y}_{i,c},
\]
where $c=0$ and $c=1$ denote the target-absent and target-containing classes, respectively.

Among the target-absent patches, those assigned high target-containing
probabilities were regarded as classifier-derived hard negatives. Let
$s_i=\hat{y}_{i,1}$ denote the predicted target-containing probability for a
negative patch $x_i\in\mathcal{P}^{-}$. In~the reported experiments, hard
negative selection was performed independently within each WSI using the
\texttt{per\_wsi} scope, rather than on a globally pooled negative-patch set.
The candidate records were ordered according to the stored within-WSI ranking,
with the predicted target-containing probability used as a secondary sorting
criterion. The~first $K$ eligible records were selected:
\[
\mathcal{H}_{K}^{-}
=
\operatorname{TopK}_{x_i \in \mathcal{P}^{-}}(s_i).
\]
Here, $K$ denotes the nominal number of selected classifier-derived negative
patches, and~$K\in\{100,300,500,1000\}$ was evaluated using the same candidate
values across all FTU categories. This common candidate set was chosen to
enable controlled \mbox{accuracy--workload} comparisons under matched sampling
budgets, rather than to define a prevalence-adaptive rule or a universally
optimal budget. When a WSI contained
fewer than $K$ eligible negative patches, all available candidates were
retained. Duplicate patch indices, missing coordinate entries, and~patches
identified as FTU-containing during coordinate checking were excluded from the
final negative-patch list. If~both the stored ranking and the predicted
probability were tied, the~original order of the merged ranking records was
retained. Therefore, the~final number of selected negative patches was at most
$K$ and could vary across WSIs. For~a cross-validation fold with training-WSI
set $\mathcal{W}_f$, the~final negative-patch count was calculated as
$\sum_{w\in\mathcal{W}_f}\min(K,N_{\mathrm{eligible},w})$ after the exclusion
rules described above. Each cross-validation fold used only the selected
negative patches belonging to its corresponding training~WSIs.

The corresponding static segmentation training set was defined~as

\[
\mathcal{T}_{\mathrm{TopK}}
=
\mathcal{P}^{+}
\cup
\mathcal{H}_{K}^{-}.
\]

This per-WSI design was used to prevent WSIs with larger tissue areas from
dominating the selected negative-patch pool and to enable matched comparisons
between classifier-guided and random~sampling.

This setting isolates the contribution of classifier-derived hard negatives and allows direct comparison with positive-only training, all-tissue training, and~random \mbox{negative~sampling.}

\subsubsection{Segmentation Network}

\begin{figure}[H]
    \centering
    \includegraphics[width=1.0\textwidth]{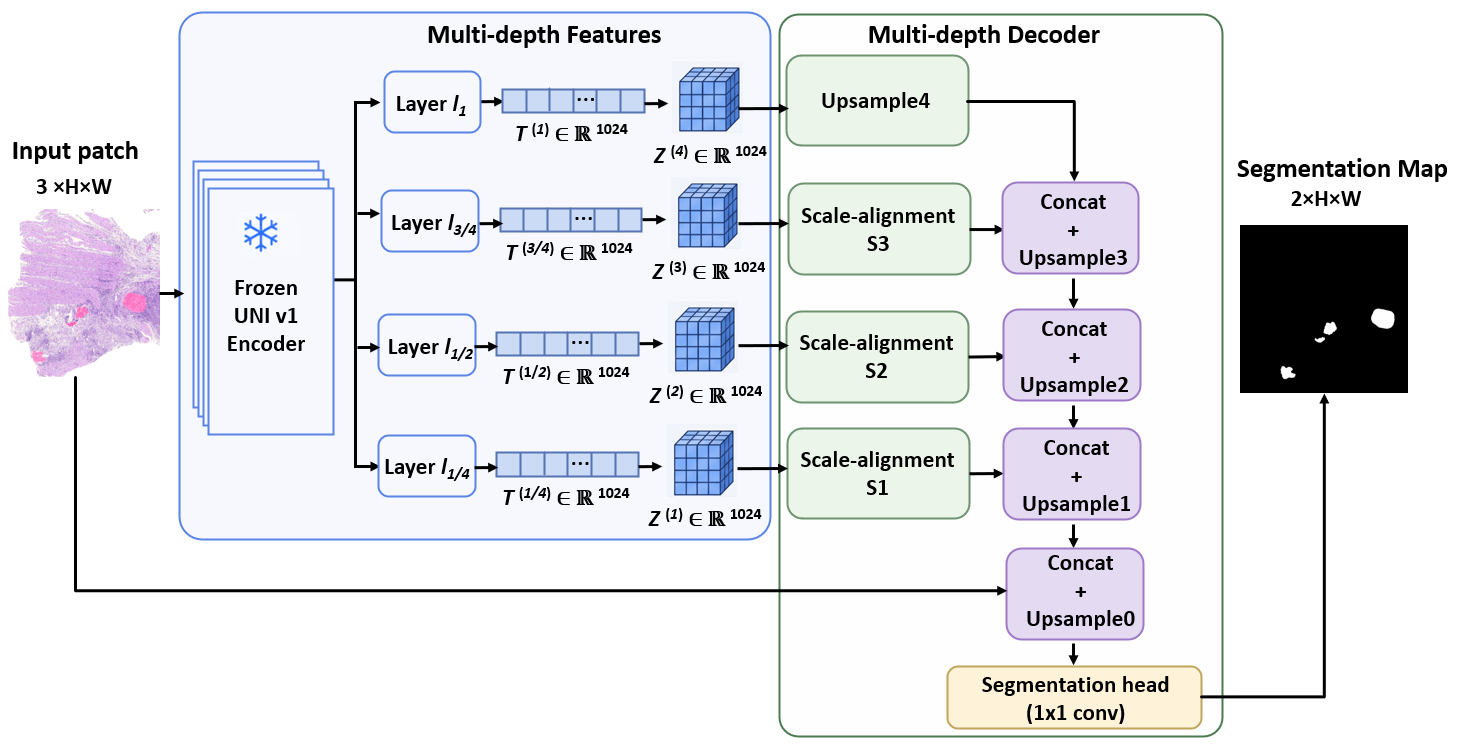}
    \caption{Architecture of the segmentation network. Four Transformer-layer features are extracted from the frozen UNI v1 encoder at relative depths of $1/4$, $1/2$, $3/4$, and~$1$, reshaped into spatial feature maps, and~progressively fused with shallow image features by a multi-scale decoder to produce dense binary segmentation~predictions.}
    \label{fig:seg_network}
\end{figure}

The segmentation network used in Stage II is illustrated in Figure~\ref{fig:seg_network}. It consists of a frozen UNI v1 encoder~\cite{chen2024towards,liu2026ftu} and a multi-scale feature fusion decoder inspired by dense prediction transformer designs~\cite{ranftl2021vision}. For~each input patch, multi-depth Transformer features extracted by UNI v1 are reshaped into spatial feature maps and progressively fused in the decoder. Starting from the deepest feature, the~decoder gradually restores spatial resolution by integrating shallower Transformer features and a shallow convolutional representation of the original image, ultimately producing a dense two-class segmentation~map.

Given an input patch $x_i \in \mathbb{R}^{3 \times H \times W}$, 
the frozen UNI v1 encoder first partitions it into non-overlapping 
$h \times w$ image tokens. This results in a token grid of size 
$H_t \times W_t$, where $H_t = H/h$ and $W_t = W/w$. 
To provide multi-level representations for dense prediction, we extract 
token features from $L$ selected relative encoder depths:
\[
T_i^{(j)}
=
E_{\mathrm{UNI}}^{(r_j)}(x_i),
\qquad
j=1,2,\ldots,L,
\]
where $E_{\mathrm{UNI}}^{(r_j)}(\cdot)$ denotes the UNI v1 output at the 
selected relative depth $r_j$. In~this study, $L=4$ and 
$r_j \in \{1/4, 1/2, 3/4, 1\}$. Each feature sequence 
contains one class 
token and $H_t \times W_t$ patch tokens, i.e.,
$T_i^{(j)}
\in
\mathbb{R}^{(1+H_t W_t)\times 1024}.$

Before decoding, the~class token is removed, and~the remaining patch-token 
subset $T_{i,\mathrm{patch}}^{(j)}$ is reshaped into a two-dimensional 
feature map:
\[
Z_i^{(j)}
=
\mathcal{R}
\left(
T_{i,\mathrm{patch}}^{(j)}
\right),
\qquad
Z_i^{(j)}
\in
\mathbb{R}^{1024 \times H_t \times W_t},
\]
where $\mathcal{R}(\cdot)$ denotes the token-to-feature-map reshaping 
operation. For~simplicity, the~four reshaped feature maps are denoted as 
$Z_1,Z_2,Z_3,Z_4$.

The decoder adopts a progressive multi-scale fusion strategy. Starting from 
the deepest feature $Z_4$, it gradually restores spatial resolution by 
integrating shallower encoder features. Taking the fusion with $Z_3$ as an 
example, $Z_4$ is first upsampled to obtain $B_4$, while $Z_3$ is transformed 
to the same spatial scale and channel dimension. The~two features are then 
concatenated and further refined:
\[
B_3
=
U_3
\left(
\operatorname{Concat}
\left(
U_4(Z_4), S_3(Z_3)
\right)
\right),
\]
where $U_4(\cdot)$ denotes the bottleneck upsampling block, $S_3(\cdot)$ 
denotes the scale-alignment and channel-projection transformation for $Z_3$, 
$U_3(\cdot)$ denotes the fusion-and-upsampling block, and~
$\operatorname{Concat}(\cdot,\cdot)$ denotes channel-wise~concatenation.

The same fusion procedure is applied to the remaining encoder features. 
Specifically, $B_3$ is fused with the transformed $Z_2$ to produce $B_2$, 
and $B_2$ is further fused with the transformed $Z_1$ to obtain $B_1$. 
In our implementation, the~decoded feature resolution is progressively 
restored from the token-grid resolution $H_t \times W_t$ to the original 
input resolution $H \times W$.

In parallel, the~original input image is processed by a shallow convolutional 
branch $S_0(\cdot)$ to preserve local texture and boundary information. 
The final segmentation prediction is obtained by concatenating the decoded 
feature $B_1$ with the shallow image feature $S_0(x_i)$ and passing the 
fused representation through the segmentation head:
\[
\hat{Y}_i
=
\operatorname{Softmax}
\left(
H
\left(
\operatorname{Concat}
\left(
B_1, S_0(x_i)
\right)
\right)
\right),
\qquad
\hat{Y}_i \in \mathbb{R}^{2 \times H \times W},
\]
where $H(\cdot)$ denotes the final convolutional prediction head, and~
$\operatorname{Concat}(\cdot,\cdot)$ denotes channel-wise concatenation. 
The two output channels correspond to the non-target and target classes, 
respectively.

During training, the~UNI v1 encoder remains frozen, while the downstream 
decoding modules and segmentation head are optimized. This design leverages 
pretrained pathology representations while allowing the task-specific 
prediction layers to adapt to the segmentation targets.
In the present experiments, the~static TopK setting was used as the primary FTU-Seek configuration. This choice keeps the training protocol reproducible and isolates the effect of classifier-guided hard-negative selection from additional segmentation-stage re-mining.
The candidate values $K\in\{100,300,500,1000\}$ were predefined exploratory operating points selected to cover progressively broader hard-negative budgets, from~highly compact training sets to more extensive negative coverage. They were not adopted from a specific prior study and were not intended to define a universally optimal or prevalence-adaptive~budget.

\subsubsection{Optimization Objectives and Training Protocol}

The patch-level classifier was optimized using cross-entropy loss. Let 
$\mathbf{y}_i=[y_{i,0},y_{i,1}]$ denote the one-hot patch-level ground-truth label, 
and let $\hat{\mathbf{y}}_i=[\hat{y}_{i,0},\hat{y}_{i,1}]$ denote the predicted class 
probabilities obtained from Equation~\eqref{eq:cls_prediction}. The~classification loss is \mbox{defined as}
\[
\mathcal{L}_{\mathrm{cls}}
=
-\frac{1}{N}
\sum_{i=1}^{N}
\sum_{c=0}^{1}
y_{i,c}\log \hat{y}_{i,c},
\]
where $N$ is the number of training patches in a mini-batch, and~
$c=0$ and $c=1$ correspond to the target-absent and target-containing classes, respectively.

To address the foreground--background imbalance inherent to sparse target segmentation, Dice--Focal loss was adopted for model optimization~\cite{sudre2017generalised,liu2025bidirectional,lin2017focal}:
$\mathcal{L}_{\mathrm{seg}}
=
\mathcal{L}_{\mathrm{Dice}}
+
\mathcal{L}_{\mathrm{Focal}}.$

Let $p_n \in [0,1]$ denote the predicted target-class probability of the $n$-th pixel and $g_n \in \{0,1\}$ denote the corresponding ground-truth label. The~Dice loss is defined as
\[
\mathcal{L}_{\mathrm{Dice}}
=
1-
\frac{
2\sum_{n=1}^{M} p_n g_n + \epsilon
}{
\sum_{n=1}^{M} p_n
+
\sum_{n=1}^{M} g_n
+
\epsilon
},
\]
where $M$ is the total number of pixels involved in the loss computation and $\epsilon$ is a small constant for numerical~stability.

The focal loss is defined as
\[
\mathcal{L}_{\mathrm{Focal}}
=
-\frac{1}{M}
\sum_{n=1}^{M}
\left[
\alpha g_n(1-p_n)^{\gamma}\log p_n
+
(1-\alpha)(1-g_n)p_n^{\gamma}\log(1-p_n)
\right],
\]
where $\alpha$ and $\gamma$ denote the balancing and focusing parameters, respectively.
Five-fold cross-validation was performed on the development set at the WSI level. In~each fold, models were trained using the corresponding training split, and~model selection was performed on the validation split. For~classifier-guided hard negative strategies, the~patch-level classifier was trained within the same fold to rank target-absent patches, and~the resulting TopK hard negatives were used to construct the segmentation training~set.

The segmentation model was trained using a Dice--Focal loss and a ReduceLROnPlateau learning rate scheduler. For~each fold, the~model with the best validation performance was retained and evaluated on the independent test~set.

\subsubsection{Comparative Patch Construction Strategies}

To assess the contribution of classifier-guided hard negative selection, we compared the following patch construction~strategies:
\begin{itemize}

    \item Positive-only training.
    This strategy trains the segmentation model only with target-containing patches:
    $\mathcal{T}_{\mathrm{pos}}=\mathcal{P}^{+}$.

    \item All-tissue training.
    This strategy uses all available tissue patches:
    $\mathcal{T}_{\mathrm{all}}=\mathcal{P}^{+}\cup\mathcal{P}^{-}$.

    \item Random negative sampling.
    
    This strategy retains all target-containing patches and randomly selects a
    target-absent subset $\mathcal{R}^{-}$ from $\mathcal{P}^{-}$. For~the matched
    random-$K$ experiments, eligible negative patches were sampled independently
    within each WSI without replacement using
    $\min(K,N_{\mathrm{eligible}})$ candidates, where
    $N_{\mathrm{eligible}}$ denotes the number of eligible negative patches
    available for that WSI. Thus, matched random-$K$ sampling used the same
    per-WSI negative-patch budget as the corresponding FTU-Seek Top$K$
    configuration, but~without classifier-based ranking. The~corresponding
    segmentation training set~was
    
    \[
    \mathcal{T}_{\mathrm{rand}}
    =
    \mathcal{P}^{+}\cup\mathcal{R}^{-}.
    \]

    \item Static FP-find TopK sampling.
    This strategy trains the model with all positive patches and the classifier-derived TopK hard negatives:
    $\mathcal{T}_{\mathrm{TopK}}=\mathcal{P}^{+}\cup\mathcal{H}_{K}^{-}$.
    This setting evaluates whether high-risk negative patches selected by the patch-level classifier are more informative than randomly selected~negatives.

\end{itemize}

\subsection{Outcomes}
The performance of the deep learning framework was evaluated across two distinct dimensions: spatial segmentation accuracy and patch-level classification performance. To~assess the spatial overlap and delineation accuracy of the segmentation network, the~Dice similarity coefficient (DSC) was utilized as the primary metric~\cite{milletari2016v,gao2022rolling}. DSC quantifies the pixel-level agreement between the model's predicted output masks and the manual ground truth annotations, providing a rigorous measure of morphological fidelity:
\[
\mathrm{DSC} = \frac{2\mathrm{TP}}{2\mathrm{TP} + \mathrm{FP} + \mathrm{FN}}.
\]

For the patch-level classification performance, a~confusion matrix was constructed to tally true positives (TPs), true negatives (TNs), false positives (FPs), and~false negatives (FNs). The~principal metric for this phase was the area under the receiver operating characteristic curve (AUC), which provides a threshold-independent measure of discriminative ability. Additionally, several secondary metrics were derived from the confusion matrix to provide a comprehensive assessment. Sensitivity (recall) and specificity were calculated to evaluate the capacity to correctly identify positive and negative cases:
\[
\mathrm{Sensitivity} = \frac{\mathrm{TP}}{\mathrm{TP} + \mathrm{FN}},\qquad
\mathrm{Specificity} = \frac{\mathrm{TN}}{\mathrm{TN} + \mathrm{FP}}.
\]

Positive predictive value (PPV) and negative predictive value (NPV) assessed the reliability of the classifier predictions:
\vspace{-6pt}
\[
\mathrm{PPV} = \frac{\mathrm{TP}}{\mathrm{TP} + \mathrm{FP}},\qquad
\mathrm{NPV} = \frac{\mathrm{TN}}{\mathrm{TN} + \mathrm{FN}}.
\]

Furthermore, overall accuracy and the F1-score---the harmonic mean of precision and sensitivity---were computed to measure holistic performance~\cite{ehteshami2017diagnostic}, particularly addressing potential class imbalance:
\[
\mathrm{Accuracy} = \frac{\mathrm{TP} + \mathrm{TN}}{\mathrm{TP} + \mathrm{TN} + \mathrm{FP} + \mathrm{FN}}, \qquad
\mathrm{F1\text{-}score} = 2 \times \frac{\mathrm{PPV} \times \mathrm{Sensitivity}}{\mathrm{PPV} + \mathrm{Sensitivity}}.
\]

For downstream analysis, segmentation outputs were converted into slide- or patient-level FTU phenotypes. TLS features included TLS area density (ratio of TLS area to total tissue area), TLS count density (number of TLS per unit area), size-stratified TLS densities (small/medium/large by area tertiles), hotspot TLS density, and~nearest-neighbor clustering indices (observed mean nearest-neighbor distance divided by random expectation). Blood-vessel features included total and filtered BV density, vessel count density, size-stratified vessel densities, and~hotspot vascular density measures. Gland features included gland density, gland count density, size-stratified gland densities, and~shape descriptors such as area, perimeter, circularity, solidity, and~eccentricity. OS was used for TLS and BV survival analyses when survival information was available. In~LIHC, the~distributions of vascular characteristics were compared between MVI-positive and MVI-negative patients. For~PRAD, glandular features were evaluated primarily against pathology-related endpoints and BCR status; BCR was analyzed as a binary endpoint because reliable BCR-specific censoring times were not available for non-recurrent cases. Additionally, we evaluated the differences in gland features across patients with high versus low Gleason scores, different pT stages, and~resection margin statuses (R0 vs. R1).

\subsection{Implementation~Details}
All model training experiments were implemented in PyTorch and performed on an NVIDIA RTX PRO 6000
 Blackwell Workstation Edition GPU with 97,887 MiB of memory. Although~two GPUs were available on the workstation, each training run was executed on a single GPU using the specified CUDA device. To~improve reproducibility, Python, NumPy, PyTorch, and~CUDA random seeds were fixed, and~deterministic cuDNN settings were~enabled.

The patch-level classifier was trained in the FP-find classification pipeline. For~each FTU task and cross-validation fold, frozen UNI v1 embeddings were extracted from ten relative encoder depths ($0.1,0.2,\ldots,1.0$). Each 1024-dimensional depth-specific feature was passed through a two-layer feed-forward transformation head with ReLU activation, expanded to 4096 dimensions and projected back to 1024 dimensions. The~transformed multi-depth features were averaged and passed to a final linear classification layer to predict target-absent versus target-containing patch labels. The~classifier was optimized using cross-entropy loss and AdamW with an initial learning rate of $5 \times 10^{-5}$, weight decay of $5 \times 10^{-6}$, batch size of 128, and~a maximum of 500 epochs. A~ReduceLROnPlateau scheduler was applied to the validation AUC, and~early stopping was used with a patience of 15 epochs. For~each fold, the~checkpoint with the highest validation AUC was selected as the validation-best classifier. This classifier was then applied to FTU-absent training patches to compute the predicted FTU-containing probability. These probabilities were used to rank candidate hard negatives, and~the top-ranked FTU-absent patches were exported for downstream static TopK segmentation~training.

The segmentation model was trained within the FTU-Seek segmentation pipeline. For~each FTU task, five-fold cross-validation was performed at the WSI level on the development set, and~the independent test set was held out for final evaluation. The~segmentation network used a frozen UNI v1 encoder and a trainable residual feature-fusion decoder. Four encoder depths ($1/4$, $1/2$, $3/4$, and~$1$) were used for dense prediction, and~the corresponding token features were reshaped into spatial feature maps before decoder fusion. Input image patches were resized to $224 \times 224$ pixels and normalized before model input. For~classifier-guided FTU-Seek sampling, the~segmentation training set contained all FTU-containing patches and the classifier-ranked FTU-absent hard negatives selected by static TopK sampling within each WSI. Matched random TopK experiments used the same number of negative patches per WSI but selected them randomly from the FTU-absent~pool.

During segmentation training, only the decoder and prediction head were updated, whereas all UNI v1 encoder parameters remained frozen. The~trainable parameters were optimized using AdamW with an initial learning rate of $5 \times 10^{-5}$, weight decay of $5 \times 10^{-6}$, batch size of 32, and~a maximum of 50 epochs. Mixed-precision training was enabled when CUDA was available using automatic casting and gradient scaling. The~segmentation objective combined multiclass Dice loss and focal loss. In~the implementation, the~focal term used class weights of 0.25 and 0.75 with a focusing parameter of $\gamma=2.0$, and~the final loss was defined as
$\mathcal{L}_{\mathrm{seg}}=\mathcal{L}_{\mathrm{Dice}}+20\mathcal{L}_{\mathrm{Focal}}$.
A ReduceLROnPlateau scheduler monitored the validation mean Dice coefficient and reduced the learning rate with a factor of 0.95, patience of 5 epochs, cooldown of 2 epochs, and~minimum learning rate of $1 \times 10^{-7}$. Validation was performed after each training epoch, and~early stopping was applied with a patience of 5~epochs. For~each fold, the~checkpoint with the highest validation mean Dice was retained and subsequently evaluated on the independent test~set.

For reproducibility, the~primary experiments were conducted using random seed
1, and~repeated random-baseline experiments used seeds 1--5. The~WSI-level
fold assignments and train--validation--test split metadata were fixed before
model training and were stored in the corresponding split files. Experiments
were run in Python 3.11.15 using PyTorch 2.8.0+cu128, torchvision 0.23.0+cu128,
CUDA 12.8, NumPy 1.26.4, pandas 2.3.3, SciPy 1.12.0, h5py 3.15.0, timm 1.0.20,
einops 0.8.1, Pillow 12.2.0, Matplotlib 3.10.7, and~seaborn 0.13.2. The~pathology foundation model was the publicly released UNI v1 model from the
Mahmood Lab.

During inference, the~foreground probability was obtained from the foreground
class of the two-class softmax output. A~fixed threshold of 0.5 was used to
convert the probability map into a binary foreground mask, with~pixels having
foreground probability $\geq 0.5$ classified as positive. Predictions outside
the foreground tissue mask were set to zero. Patch-level predictions were mapped back to their original WSI coordinates
for slide-level mask visualization and downstream tissue phenotyping. For~Dice
calculation, pixel-level true-positive, false-positive, and~false-negative
counts were accumulated across all evaluated patches within each WSI. Thus,
the reported Dice was not obtained by averaging patch-level Dice values.
No connected-component filtering, hole filling, minimum-instance-size filtering,
or other morphological post-processing was applied before Dice~calculation.

\subsection{Statistical~Analysis}

Segmentation and classification metrics on the development set and internal
test set were summarized as mean $\pm$ standard deviation across five
cross-validation folds.
For the multi-seed experiments, the~original run used seed 1, and~four additional random-baseline runs using seeds 2--5 were subsequently performed. Random-baseline results were summarized across all five seeds (1--5) using the mean and standard deviation. These standard deviations quantify variability caused by random sampling and stochastic training across seeds. Because~the original FP-find experiments were conducted using seed 1, the~multi-seed analysis was interpreted primarily as an assessment of the stability of the random-sampling baselines rather than as a direct estimate of FP-find seed~robustness.

For the 30-slide held-out TLS evaluation, Dice scores were calculated
independently for each WSI and summarized as mean $\pm$ standard deviation
across slides. For~each paired comparison, the~paired difference was
calculated as the Dice score of FTU-Seek minus that of the comparator strategy.
Two-sided 95\% confidence intervals for the mean paired differences were
calculated using the Student's $t$ distribution with 29 degrees of freedom.
Paired $t$-tests and Wilcoxon signed-rank tests were performed using the WSI
as the pairing unit. The~comparisons included matched random Top$K$ strategies
at $K=100$, $300$, $500$, and~$1000$, as~well as the representative
Top$1000$ FTU-Seek strategy versus all-tissue and random-balanced training.
All formal paired analyses were restricted to the independent 30-WSI held-out
TLS cohort. The~internal test cohort contained only two~WSIs and was therefore
retained for descriptive comparison rather than formal hypothesis testing.
Predictions for the held-out cohort were generated using five-fold probability
ensembles, in~which foreground probabilities from the five-fold models were
averaged before thresholding. The~reported $p$-values were nominal two-sided
values.

For segmentation, the~Dice coefficient was reported on both validation folds and independent test sets, and~training workload was calculated as the proportion of retained training patches relative to all-tissue training within each~cohort.

For each evaluation WSI, pixel-level predictions and reference masks were
processed patch by patch, while the numbers of true-positive, false-positive,
and false-negative pixels were accumulated across all evaluated patches from
that WSI. The~WSI-level Dice coefficient was then calculated as
\[
\mathrm{Dice}_{\mathrm{WSI}}
=
\frac{2TP_{\mathrm{WSI}}}
{2TP_{\mathrm{WSI}}+FP_{\mathrm{WSI}}+FN_{\mathrm{WSI}}}.
\]
Thus, the~reported Dice was not obtained by averaging patch-level Dice
coefficients and was not an instance-level metric. Because~the extracted
patches were non-overlapping, this accumulation corresponds to pixel-level
evaluation over the complete WSI tissue region without double-counting
overlapping~patches.

For the original development and internal-test experiments, Dice values were
summarized across the five cross-validation folds. For~the held-out TLS
analysis, one WSI-level Dice value was obtained for each of the 30 WSIs, and~these WSI-level values were used for the paired comparisons and confidence
interval~calculations.

For downstream survival analyses, OS was defined as the interval from diagnosis or surgery to death from any cause, with~surviving patients censored at last follow-up. A~total of 60 TLS features were evaluated separately in each of TCGA-READ, TCGA-ESCA, and~TCGA-STAD, while 36 vascular features were evaluated in TCGA-LIHC. For~TLS analyses, pairwise Wilcoxon rank-sum tests were used for descriptive cross-cohort comparison of TLS morphology and spatial organization, and~associations with OS were evaluated using Cox proportional hazards models with individual TLS features entered as continuous variables. Features were standardized to facilitate comparison across different measurement scales, and~hazard ratios (HRs) with 95\% confidence intervals (CIs) were reported per 1-standard-deviation (SD) increase. To~account for multiple feature-wise survival analyses, the~Benjamini--Hochberg false discovery rate (FDR) procedure was applied separately within each TLS cohort across the 60 Cox regression tests performed in that cohort. Kaplan--Meier analyses were used as exploratory visualizations of selected TLS phenotypes. For~exploratory cut-point analysis, candidate thresholds were searched under minimum group-size constraints to avoid extreme boundary splits, and~the threshold with the smallest log-rank $p$ value was retained for~visualization.

For LIHC, vascular phenotypes were further examined in relation to OS and MVI. For~survival analysis, selected features were dichotomized using the exploratory data-derived cut points described above and evaluated using Cox proportional hazards models, with~HRs and 95\% CIs reported for the high- versus low-feature groups. Because~these cut points were outcome-derived, the~corresponding survival analyses were considered exploratory and were not interpreted as defining validated prognostic thresholds. Multivariable Cox regression was used to assess whether the vascular phenotype remained associated with OS after adjustment for available clinical covariates. The~proportional-hazards assumption was assessed using a feature-by-log-transformed-time interaction term, and~multicollinearity among predictors was evaluated using variance inflation factors (VIFs). For~MVI, all 36 vascular features were compared between patients with and without MVI using the Wilcoxon rank-sum test and summarized as median and interquartile range (IQR), with~rank-biserial correlation reported as the effect size. The~Benjamini--Hochberg FDR procedure was applied across the 36 feature-wise MVI comparisons within TCGA-LIHC. Features of interest were subsequently evaluated in multivariable logistic regression as continuous variables, with~associations reported as adjusted odds ratios (ORs) and corresponding 95\% CIs. Available clinical covariates were included without univariable statistical pre-screening, except~for tumor stage, which was excluded because of its close relationship with vascular invasion and the potential for information overlap. No missing data were present among the 337 patients included in the analysis. Continuous vascular predictors were Z-standardized before regression, and~associations were reported as adjusted ORs with 95\% CIs per 1-SD increase. Multicollinearity was assessed using VIFs, and~potential nonlinearity of the vascular phenotype--MVI association was examined using restricted cubic splines (RCSs). The~multivariable regression analyses were used to assess adjusted associations rather than to develop clinical prediction~models.

Similarly, for~PRAD glandular analyses, 40 glandular features were evaluated according to Gleason score ($\leq$7 vs. $>$7), pathological T stage (T2 vs. T3/T4), surgical margin status (R0 vs. R1), and~BCR status (No vs. Yes). Continuous glandular features were compared between groups using the Wilcoxon test and summarized as median and interquartile range (IQR). Effect sizes were reported using rank-biserial correlation. The~Benjamini--Hochberg FDR procedure was applied separately within each clinicopathological endpoint, corresponding to 40 feature-wise comparisons for Gleason score, 40 for pathological T stage, 40 for surgical margin status, and~40 for BCR status. All statistical tests were two-sided, and~$p < 0.05$ was considered nominally significant. For~the exploratory TLS survival analyses and the vascular and glandular feature-comparison analyses, FDR-adjusted $p < 0.10$ was used as an exploratory multiple-testing threshold. The~downstream analyses were designed as proof-of-concept evaluations of associations between segmentation-derived phenotypes and clinicopathological characteristics rather than as confirmatory biomarker-validation analyses, and~the resulting clinical associations were therefore considered~hypothesis-generating.

\section{Results}

\subsection{Dataset~Characteristics}
As shown in \Cref{tab:dataset_characteristics}, the~three FTU segmentation cohorts differed substantially in target abundance and spatial sparsity at the instance, patch, and~pixel levels. At~the instance level, the~development sets contained 133 TLSs, 2508 blood vessels, and~3074 glands, corresponding to mean numbers of 16.62, 358.29, and~384.25 instances per WSI, respectively. A~similar pattern was observed in the test sets, indicating that TLSs were considerably less frequent than blood vessels and~glands.

\begin{table}[htbp]
\caption{Dataset characteristics and spatial sparsity of the three FTU segmentation cohorts at instance, patch, and pixel levels.}
\label{tab:dataset_characteristics}
\begin{tabular*}{\textwidth}{@{\extracolsep{\fill}}lccc@{}}
\toprule
\textbf{Characteristics} & \textbf{TLS} & \textbf{BV} & \textbf{Gland} \\
\midrule
\multicolumn{4}{l}{\textit{Development Set}} \\
\multicolumn{4}{l}{Instance level} \\
Unique patients & 8 & 7 & 8 \\
WSIs & 8 & 7 & 8 \\
WSIs per patient & 1 & 1 & 1 \\
Total FTU instances & 133 & 2508 & 3074 \\
Mean FTU instances per WSI & 16.62 & 358.29 & 384.25 \\
Median FTU instances per WSI & 12.50 & 316.00 & 377.50 \\
\multicolumn{4}{l}{Patch level} \\
Total tissue patches & 33,752 & 20,831 & 26,366 \\
FTU-containing patches & 1322 & 3422 & 9719 \\
FTU-absent patches & 32,430 & 17,409 & 16,647 \\
FTU-containing patch ratio (\%) & 3.92 & 16.43 & 36.86 \\
\multicolumn{4}{l}{Pixel level} \\
Tissue pixel area & 145,723,284 & 67,053,366 & 80,370,006 \\
FTU pixel area & 942,758 & 1,646,695 & 15,197,324 \\
FTU pixel area ratio (\%) & 0.65 & 2.46 & 18.91 \\
\midrule
\multicolumn{4}{l}{\textit{Test Set}} \\
\multicolumn{4}{l}{Instance level} \\
Unique patients & 2 & 2 & 2 \\
WSIs & 2 & 2 & 2 \\
WSIs per patient & 1 & 1 & 1 \\
Total FTU instances & 105 & 409 & 453 \\
Mean FTU instances per WSI & 52.50 & 204.50 & 226.50 \\
Median FTU instances per WSI & 52.50 & 204.50 & 226.50 \\
\multicolumn{4}{l}{Patch level} \\
Total tissue patches & 9421 & 5222 & 5432 \\
FTU-containing patches & 757 & 583 & 1951 \\
FTU-absent patches & 8664 & 4639 & 3481 \\
FTU-containing patch ratio (\%) & 8.04 & 11.16 & 35.92 \\
\multicolumn{4}{l}{Pixel level} \\
Tissue pixel area & 38,848,017 & 16,370,435 & 16,686,067 \\
FTU pixel area & 901,489 & 273,117 & 3,981,224 \\
FTU pixel area ratio (\%) & 2.32 & 1.67 & 23.86 \\
\bottomrule
\end{tabular*}

\vspace{1ex}
\noindent{\footnotesize{Abbreviations: FTU, functional tissue unit; TLS, tertiary lymphoid structure; BV, blood vessel; WSI, whole-slide image. Dataset statistics were calculated from the actual slides included in the train\_val\_test.json split files used for segmentation experiments. Development and test set statistics are reported separately in this table and were not pooled. ``FTU-containing patches'' denotes tissue patches overlapping annotated FTU regions, whereas ``FTU-absent patches'' denotes tissue patches without annotated FTU regions. Patch-level ratios were calculated as FTU-containing patches divided by total tissue patches. Pixel-level ratios were calculated as FTU pixel area divided by total tissue pixel area.}}
\end{table}

The patch-level statistics further demonstrated pronounced class imbalance. In~the TLS development set, only 1322 of 33,752 tissue patches contained annotated TLS regions, yielding an FTU-containing patch ratio of 3.92\%. The~corresponding ratio increased to 8.04\% in the test set, with~757 positive patches among 9421 tissue patches. For~BV segmentation, 3422 of 20,831 development patches and 583 of 5222 test patches contained target regions, corresponding to ratios of 16.43\% and 11.16\%, respectively. In~contrast, glands were more densely distributed, with~FTU-containing patch ratios of 36.86\% in the development set and 35.92\% in the test~set.

This difference was also evident at the pixel level. TLS regions accounted for only 0.65\% of the tissue area in the development set and 2.32\% in the test set. BV pixels represented 2.46\% and 1.67\% of the tissue area, respectively, whereas gland regions occupied substantially larger proportions of 18.91\% and 23.86\%. Collectively, these statistics show that TLS segmentation represents the most spatially sparse and imbalanced task, BV segmentation exhibits moderate sparsity, and~gland segmentation contains comparatively abundant target regions at both patch and pixel~levels.

As summarized in \Cref{tab:tcga_baseline,tab:tcga_downstream_cohorts}, external TCGA cohorts were used for exploratory downstream application across the three FTU segmentation tasks. For~TLS, OS analyses were conducted in the TCGA-READ, TCGA-ESCA, and~TCGA-STAD cohorts. For~BV, vascular phenotypes in TCGA-LIHC were evaluated in relation to OS and MVI. For~gland segmentation, the~TCGA-PRAD cohort was evaluated against multiple clinicopathological endpoints, including Gleason score, pathological T stage, surgical margin status, and~BCR status. These cohorts enabled an exploratory assessment of whether the automatically quantified FTU features were associated with patient prognosis and clinically relevant pathological characteristics across different cancer~types.

\begin{table}[htbp]
\caption{Baseline characteristics of the TCGA cohorts.}
\label{tab:tcga_baseline}
\small
\begin{tabular*}{\textwidth}{@{\extracolsep{\fill}}llllll@{}}
\toprule
 & \textbf{TCGA-} & \textbf{TCGA-} & \textbf{TCGA-} & \textbf{TCGA-} & \textbf{TCGA-} \\
 & \textbf{STAD} & \textbf{READ} & \textbf{ESCA} & \textbf{LIHC} & \textbf{PRAD} \\
 & \boldmath{$(n = 326)$} & \boldmath{$(n = 146)$} & \boldmath{$(n = 148)$} & \boldmath{$(n = 337)$} & \boldmath{$(n = 401)$} \\ \midrule
Sex & & & & & \\
\quad Male & 215 (65.95) & 82 (56.16) & 128 (86.49) & 228 (67.66) & 401 (100) \\
\quad Female & 111 (34.05) & 64 (43.84) & 20 (13.51) & 109 (32.34) & 0 (0) \\
Age & 66 (57--72) & 65 (57--72) & 59 (53--69) & 61 (51--68) & 61 (56--66) \\
Stage & & & & & \\
\quad I & 39 (11.96) & 28 (19.18) & 34 (22.97) & 166 (49.26) & 29 (7.23) \\
\quad II & 109 (33.44) & 46 (31.51) & 49 (33.11) & 81 (24.04) & 121 (30.17) \\
\quad III & 159 (48.77) & 51 (35.62) & 47 (31.76) & 83 (24.63) & 192 (47.88) \\
\quad IV & 19 (5.83) & 21 (13.70) & 18 (12.16) & 7 (2.08) & 59 (14.71) \\
Median Follow-up & 26.68 & 25.03 & 20.01 & 28.88 & 34.83 \\
\quad (months, 95\% CI) & (22.67--29.57) & (20.53--30.98) & (14.85--25.20) & (25.07--35.06) & (31.67--37.95) \\
Status & & & & & \\
\quad Alive & 194 (59.51) & 120 (82.19) & 86 (58.11) & 216 (64.09) & 391 (97.51) \\
\quad Dead & 132 (40.49) & 26 (17.81) & 62 (41.89) & 121 (35.91) & 10 (2.49) \\
Survival Rate (\%) & & & & & \\
\quad 1-year & 79.52 & 93.19 & 76.87 & 84.92 & 99.73 \\
\quad 3-year & 50.22 & 82.86 & 40.57 & 63.73 & 98.26 \\
\quad 5-year & 37.24 & 49.60 & 19.13 & 49.67 & 97.61 \\
\bottomrule
\end{tabular*}

\vspace{1ex}
\noindent{\footnotesize{Data are n (\%) or median (IQR). TCGA, The Cancer Genome Atlas. Stage is based on the 8th edition of AJCC staging.}}
\end{table}

\begin{table}[htbp]
\footnotesize
\caption{External TCGA cohorts used for exploratory downstream application.}
\label{tab:tcga_downstream_cohorts}
\begin{tabular*}{\textwidth}{@{\extracolsep{\fill}}lllll@{}}
\toprule
\textbf{FTU Task} & \textbf{TCGA Cohort} & \textbf{Analysis Endpoint} & \textbf{Matched Patients} & \textbf{Events/Groups} \\
\midrule
TLS & READ & Overall survival & 146 & 26 deaths \\
TLS & ESCA & Overall survival & 148 & 62 deaths \\
TLS & STAD & Overall survival & 326 & 132 deaths \\
BV & LIHC & Overall survival & 337 & 121 deaths \\
BV & LIHC & MVI & 337 & 100 positive vs. 237 negative \\
Gland & PRAD & Gleason score & 401 & 151 high vs. 250 low \\
Gland & PRAD & Pathological T stage & 395 & 239 T3/T4 vs. 156 T2 \\
Gland & PRAD & Surgical margin & 372 & 115 R1 vs. 257 R0 \\
Gland & PRAD & BCR status & 342 & 301 non-recurrent \\
\bottomrule
\end{tabular*}

\vspace{1ex}
\noindent{\footnotesize{Abbreviations: BCR, biochemical recurrence; BV, blood vessel; ESCA, esophageal carcinoma; LIHC, liver hepatocellular carcinoma; FTU, functional tissue unit; PRAD, prostate adenocarcinoma; READ, rectum adenocarcinoma; STAD, stomach adenocarcinoma; TCGA, The Cancer Genome Atlas; TLS, tertiary lymphoid structure.}}
\end{table}

\subsection{Patch-Level Classifier~Performance}

Before segmentation training, we evaluated whether the patch-level classifier could distinguish FTU-containing patches from FTU-absent tissue patches and provide a reliable ranking score for hard-negative~selection.

\begin{figure}[H]
    \includegraphics[width=1.0\columnwidth]{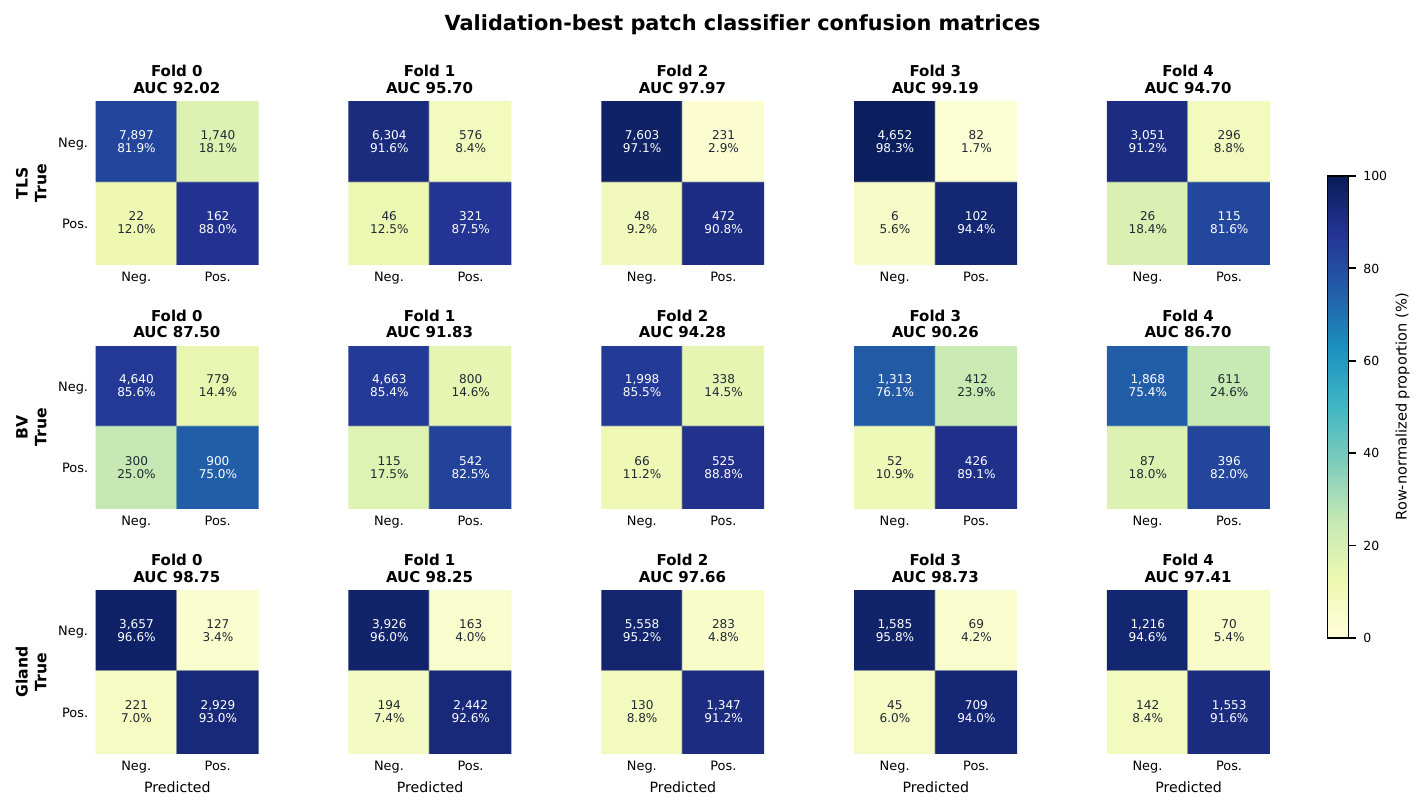}
\caption{Confusion matrices detailing the patch-level classification performance of FTUs across 5-fold cross-validation. The~confusion matrices illustrate the validation-best prediction results for three independent FTU segmentation cohorts: tertiary lymphoid structure (TLS, top row), blood vessel (BV, middle row), and~gland (bottom row). Columns correspond to the validation set results from Fold 0 through Fold 4, demonstrating the robust generalization of the deep learning framework. Within~each matrix, the~absolute counts of image patches are presented alongside the row-normalized proportions, detailing the true positive rate (sensitivity) and true negative rate (specificity). The~color intensity of the heatmap corresponds to these row-normalized percentages. The~overall area under the receiver operating characteristic curve (AUC) achieved for each specific fold and classification task is annotated above the respective~subplot.}
    \label{fig:patch_classifier_best_confusion_heatmap_grid}
\end{figure}

As summarized in \Cref{tab:patch_classifier} and visualized in \Cref{fig:patch_classifier_best_confusion_heatmap_grid}, the~validation-best patch classifiers achieved consistently strong discrimination across the three FTU tasks, although~their error profiles differed according to target prevalence and morphological complexity. The~TLS classifier achieved a mean AUC of $95.92 \pm 2.81\%$, with~a sensitivity of $88.46 \pm 4.74\%$ and a specificity of $92.01 \pm 6.46\%$. The~corresponding confusion matrices showed true-positive rates ranging from 81.6\% to 94.4\% and true-negative rates ranging from 81.9\% to 98.3\%. Its high NPV of $99.48 \pm 0.30\%$ indicates reliable exclusion of TLS-absent patches, whereas the lower and more variable PPV of $38.97 \pm 23.04\%$ reflects the pronounced rarity of TLS-containing patches and the resulting sensitivity of precision to class~prevalence.

\begin{table}[htbp]
\caption{Validation-best patch-level classifier performance across five-fold cross-validation. AUC is the primary metric. Values are percentages.}
\label{tab:patch_classifier}
\footnotesize 
\begin{tabular*}{\textwidth}{@{\extracolsep{\fill}}llcccccc@{}}
\toprule
\textbf{Cohort} & \textbf{Fold} & \textbf{AUC} & \textbf{Sensitivity} & \textbf{Specificity} & \textbf{FPR} & \textbf{NPV} & \textbf{PPV} \\
\midrule
\multirow{6}{*}{TLS}
& 0 & 92.02 & 88.04 & 81.94 & 18.06 & 99.72 & 8.52 \\
& 1 & 95.70 & 87.47 & 91.63 & 8.37 & 99.28 & 35.79 \\
& 2 & 97.97 & 90.77 & 97.05 & 2.95 & 99.37 & 67.14 \\
& 3 & 99.19 & 94.44 & 98.27 & 1.73 & 99.87 & 55.43 \\
& 4 & 94.70 & 81.56 & 91.16 & 8.84 & 99.16 & 27.98 \\
& Mean $\pm$ SD & 95.92 $\pm$ 2.81 & 88.46 $\pm$ 4.74 & 92.01 $\pm$ 6.46 & 7.99 $\pm$ 6.46 & 99.48 $\pm$ 0.30 & 38.97 $\pm$ 23.04 \\
\midrule
\multirow{6}{*}{BV}
& 0 & 87.50 & 75.00 & 85.62 & 14.38 & 93.93 & 53.60 \\
& 1 & 91.83 & 82.50 & 85.36 & 14.64 & 97.59 & 40.39 \\
& 2 & 94.28 & 88.83 & 85.53 & 14.47 & 96.80 & 60.83 \\
& 3 & 90.26 & 89.12 & 76.12 & 23.88 & 96.19 & 50.84 \\
& 4 & 86.70 & 81.99 & 75.35 & 24.65 & 95.55 & 39.32 \\
& Mean $\pm$ SD & 90.11 $\pm$ 3.11 & 83.49 $\pm$ 5.82 & 81.60 $\pm$ 5.36 & 18.40 $\pm$ 5.36 & 96.01 $\pm$ 1.39 & 49.00 $\pm$ 9.11 \\
\midrule
\multirow{6}{*}{Gland}
& 0 & 98.75 & 92.98 & 96.64 & 3.36 & 94.30 & 95.84 \\
& 1 & 98.25 & 92.64 & 96.01 & 3.99 & 95.29 & 93.74 \\
& 2 & 97.66 & 91.20 & 95.15 & 4.85 & 97.71 & 82.64 \\
& 3 & 98.73 & 94.03 & 95.83 & 4.17 & 97.24 & 91.13 \\
& 4 & 97.41 & 91.62 & 94.56 & 5.44 & 89.54 & 95.69 \\
& Mean $\pm$ SD & 98.16 $\pm$ 0.61 & 92.49 $\pm$ 1.12 & 95.64 $\pm$ 0.80 & 4.36 $\pm$ 0.80 & 94.82 $\pm$ 3.26 & 91.81 $\pm$ 5.47 \\
\bottomrule
\end{tabular*}
\end{table}

The BV classifier achieved a mean AUC of $90.11 \pm 3.11\%$, with~a sensitivity of $83.49 \pm 5.82\%$ and a specificity of $81.60 \pm 5.36\%$. Across the five folds, true-positive rates ranged from 75.0\% to 89.1\%, while true-negative rates ranged from 75.4\% to 85.6\%. The~relatively high mean FPR of $18.40 \pm 5.36\%$, particularly in Folds~3 and~4, indicates that nonvascular tissue structures with vessel-like morphology were more frequently classified as positive. Nevertheless, the~classifier maintained a high NPV of $96.01 \pm 1.39\%$, supporting its utility for excluding patches unlikely to contain vascular~structures.

Among the three tasks, the~gland classifier demonstrated the strongest and most stable performance, achieving a mean AUC of $98.16 \pm 0.61\%$, sensitivity of $92.49 \pm 1.12\%$, and~specificity of $95.64 \pm 0.80\%$. The~confusion matrices showed consistently high true-positive rates of 91.2--94.0\% and true-negative rates of 94.6--96.6\%, together with a low mean FPR of $4.36 \pm 0.80\%$. Its PPV and NPV reached $91.81 \pm 5.47\%$ and $94.82 \pm 3.26\%$, respectively, indicating reliable identification of both gland-containing and gland-absent patches. Collectively, these results demonstrate that the patch classifiers provided effective initial screening for all three FTU segmentation tasks, with~gland classification showing the greatest stability, TLS classification being primarily affected by extreme target sparsity, and~BV classification exhibiting greater susceptibility to morphologically similar background~regions.

\begin{figure}[H]
    \centering
    \includegraphics[width=1.0\columnwidth]{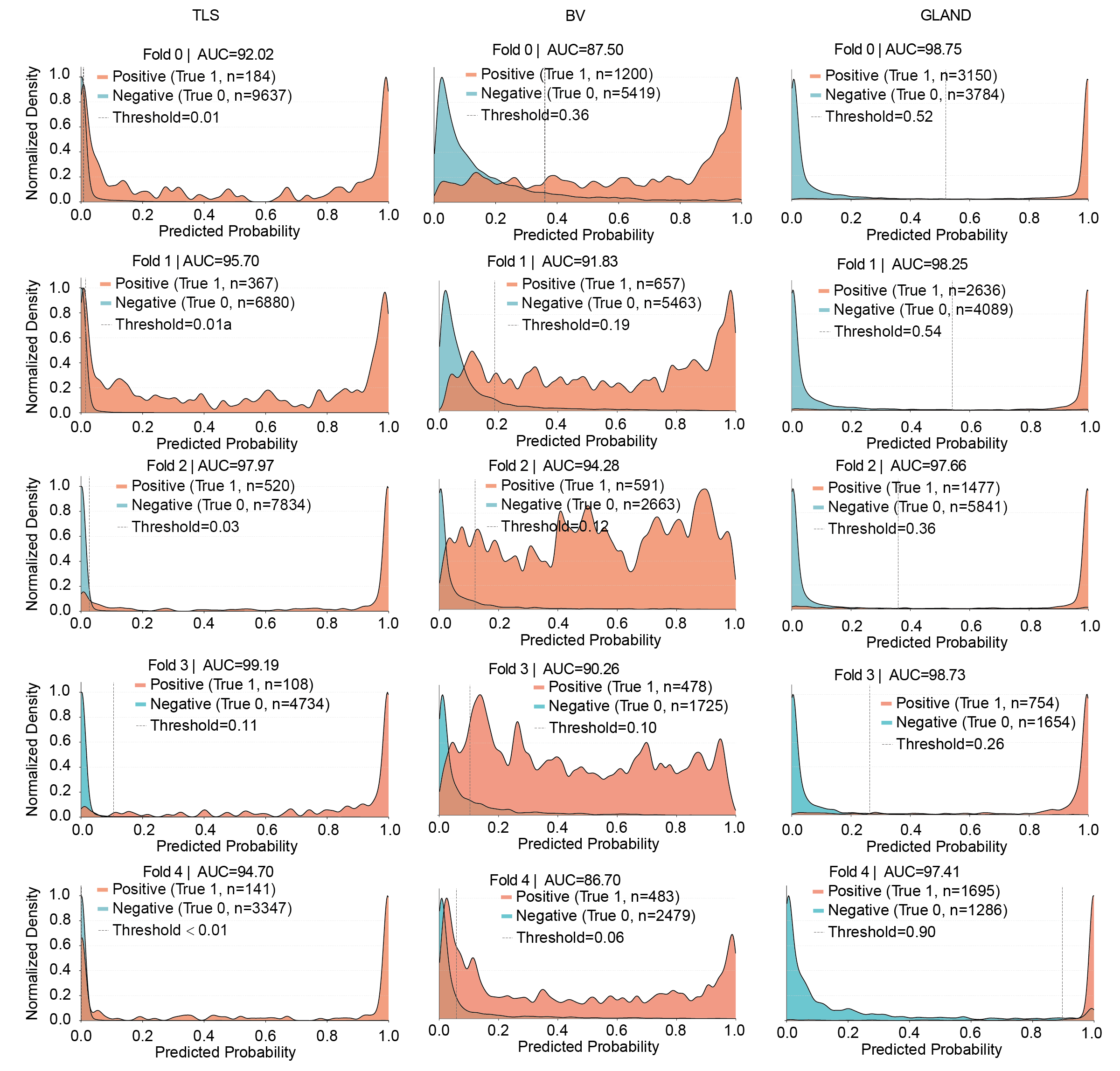}
    \caption{Validation-best predicted probability distributions of the patch-level classifiers across five-fold cross-validation. Columns correspond to the TLS, BV, and~gland cohorts, whereas rows represent Fold 0 through Fold 4. Each panel shows the normalized density distributions of the predicted FTU-containing probabilities for annotated FTU-containing patches (orange; True 1) and FTU-absent patches (teal; True 0); \(n\) indicates the number of patches in each class. The~black dashed vertical line indicates the fold-specific decision threshold, whose numerical value is provided within each panel. The~validation AUC obtained from the best-performing checkpoint is reported in the corresponding panel title.
    }
    \label{fig:patch_classifier_probability_distributions}
\end{figure}

As shown in \Cref{fig:patch_classifier_probability_distributions}, the~predicted probability distributions revealed distinct task-specific patterns of class separability across the five validation folds. For~TLS, FTU-absent patches were predominantly concentrated near a predicted probability of 0, whereas TLS-containing patches generally accumulated near 1.0. This pronounced bimodal separation was particularly evident in Folds~2 and~3, which achieved AUCs of 97.97\% and 99.19\%, respectively. Nevertheless, a~subset of TLS-containing patches received low predicted probabilities in several folds, resulting in relatively low fold-specific decision thresholds ranging from below 0.01 to 0.11. This pattern is consistent with the extreme rarity and heterogeneous appearance of TLS-containing~patches.

The BV classifiers exhibited substantially greater overlap between the positive and negative probability distributions. Although~most BV-absent patches remained concentrated toward low probabilities, BV-containing patches were broadly distributed across the full probability range, particularly in Folds~2--4. The~fold-specific thresholds consequently varied from 0.06 to 0.36. This less distinct separation agrees with the lower AUC and higher false-positive rates observed for BV classification, suggesting that vessel-containing patches were more difficult to distinguish from morphologically similar background~tissue.

In contrast, the~gland classifiers demonstrated the clearest and most consistent separation between classes. Across all folds, gland-absent patches were sharply concentrated near 0, while gland-containing patches formed prominent peaks near 1.0, with~only limited overlap between the two distributions. This pattern was reflected in the consistently high fold-wise AUCs of 97.41--98.75\%. Although~the selected decision thresholds varied from 0.26 to 0.90, the~strong separation of the two distributions resulted in stable sensitivity and specificity across folds. Overall, the~probability distributions corroborate the quantitative results, demonstrating strong class separability for gland and TLS classification, while highlighting the greater ambiguity of the BV~task.

\begin{figure}[H]
    \centering
    \includegraphics[width=1.0\columnwidth]{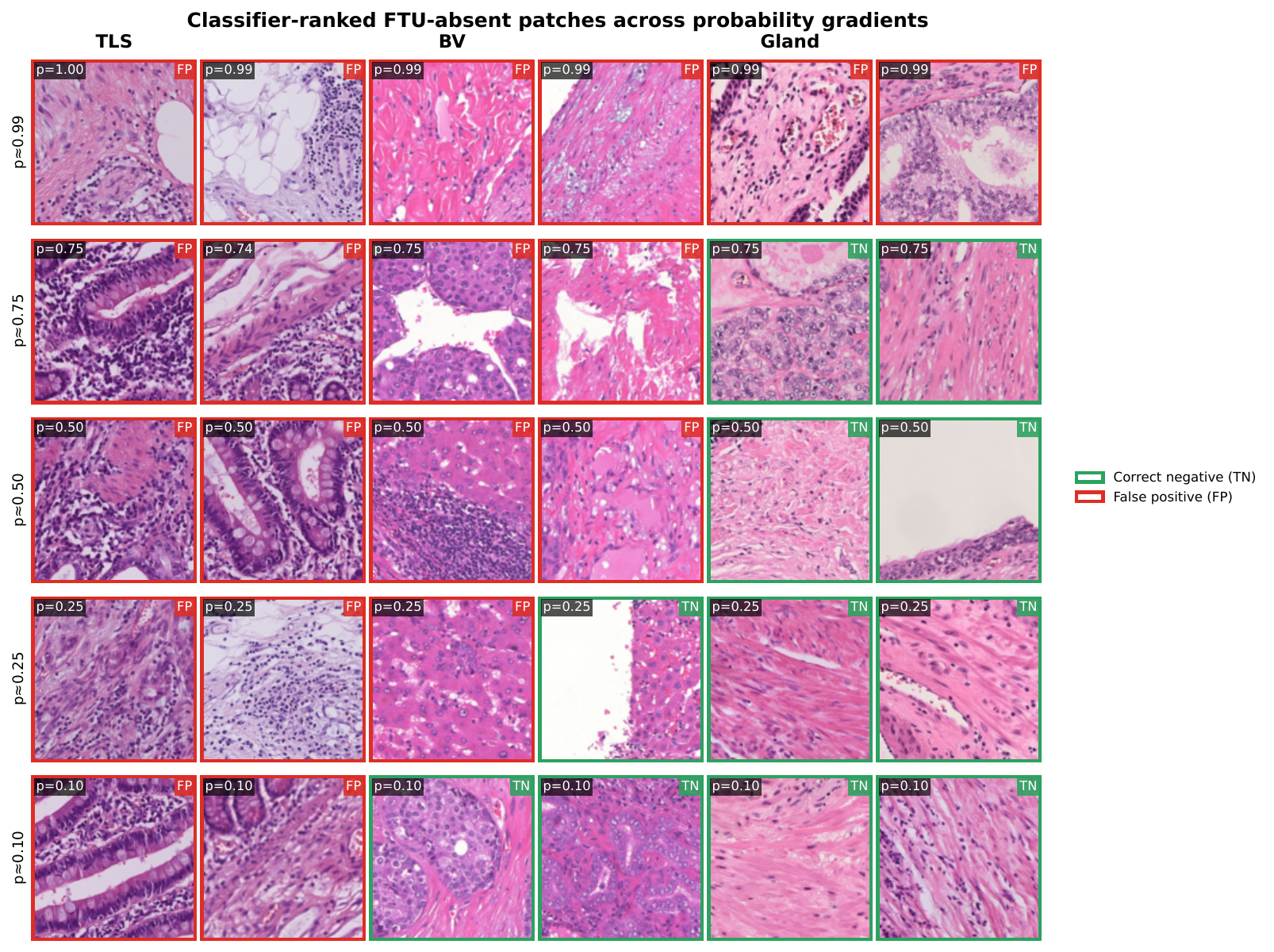}
    \caption{Representative classifier-ranked FTU-absent patches across predicted probability gradients. For~each cohort, two FTU-absent tissue patches are shown at each approximate predicted FTU-containing probability level. Patch borders indicate classification correctness according to the validation-best cutoff of the corresponding fold: green denotes correctly classified negative patches (TN), whereas red denotes false-positive patches (FP) misclassified as FTU-containing. The~examples illustrate how the patch-level classifier assigns high probabilities to morphologically confusing background regions, which are subsequently used as hard negatives for segmentation~training.}
    \label{fig:hard_negative_probability_gradient}
\end{figure}

Representative hard-negative examples are shown in \Cref{fig:hard_negative_probability_gradient}. TLS hard negatives often contained dense inflammatory or stromal regions adjacent to adipose or gland-like spaces, BV hard negatives included lumen-like or collagen-rich structures, and~gland hard negatives contained epithelial or stromal patterns resembling gland boundaries. These examples support the use of classifier-derived probability scores to prioritize FTU-absent patches that are visually similar to target-containing~regions.

\subsection{Segmentation Performance and Training-Data~Workload}

\Cref{tab:tls_all_evaluation_sets,tab:representative_bv_gland,tab:budget_bv_gland} summarize
 the segmentation accuracy and training workload of different patch construction strategies. Overall, positive-only training produced the weakest or near-weakest performance, indicating that FTU-absent tissue is necessary for suppressing false-positive predictions. All-tissue training provided the strongest reference performance in most settings but required the complete tissue-patch set. In~contrast, classifier-guided FP-find Top$K$ sampling retained only a fraction of the training data while preserving much of the independent-test~performance.

\begin{table}[htbp]
\caption{TLS Dice performance (\%) across hard-negative budgets and evaluation~cohorts.}
\label{tab:tls_all_evaluation_sets}
\small 
\begin{tabular*}{\textwidth}{@{\extracolsep{\fill}}ccccccc@{}}
\toprule
& \multicolumn{2}{c}{\begin{tabular}[c]{@{}c@{}}\textbf{Development-Set}\\ \textbf{Validation}\end{tabular}}
& \multicolumn{2}{c}{\textbf{Internal Test}}
& \multicolumn{2}{c}{\begin{tabular}[c]{@{}c@{}}\textbf{Held-Out}\\ \textbf{TLS Cohort}\end{tabular}} \\
\cmidrule(lr){2-3} \cmidrule(lr){4-5} \cmidrule(lr){6-7} 
\boldmath{$K$}
& \textbf{FTU-Seek} & \begin{tabular}[c]{@{}c@{}}\textbf{Random}\\ \textbf{(5 Seeds)}\end{tabular}
& \textbf{FTU-Seek} & \begin{tabular}[c]{@{}c@{}}\textbf{Random}\\ \textbf{(5 Seeds)}\end{tabular}
& \textbf{FTU-Seek} & \begin{tabular}[c]{@{}c@{}}\textbf{Random}\\ \textbf{(5 Seeds)}\end{tabular} \\
\midrule
100
& 68.75 & $66.62 \pm 3.03$
& 85.08 & $79.86 \pm 3.45$
& 73.99 & $68.87 \pm 1.95$ \\
300
& 73.64 & $68.82 \pm 1.63$
& 84.69 & $80.88 \pm 1.55$
& 76.54 & $69.92 \pm 2.74$ \\
500
& 72.63 & $70.28 \pm 0.79$
& 85.14 & $81.66 \pm 1.24$
& 75.19 & $70.33 \pm 2.00$ \\
1000
& 76.91 & $73.54 \pm 2.25$
& 85.12 & $83.92 \pm 0.54$
& 76.69 & $73.48 \pm 0.62$ \\
\bottomrule
\end{tabular*}

\vspace{1ex}
\noindent{\footnotesize{For each random seed, the~model was trained using five-fold
cross-validation. The~fold-level performance estimates were first
averaged within each seed to obtain a seed-level mean, and~the final
Random value was then calculated as the mean across five random seeds.
Thus, the~Random entries summarize performance hierarchically across
five folds and five seeds.
For the compact budget comparison, FTU-Seek entries correspond to the
original seed-1 experiments and are reported as fold-mean performance;
the fold-level variability of the representative Top1000 configuration
is reported in Table~\ref{tab:tls_representative_strategies}.
For the held-out TLS cohort, each fold-specific model was evaluated on
the same 30 held-out WSIs, and~the resulting performance estimates were
summarized within folds, then across folds within each seed, and~finally
across the five random seeds. These values therefore represent aggregate
performance summaries and are distinct from the WSI-level paired
analyses reported separately in Table~\ref{tab:tls_heldout_paired}.}}
\end{table}

\begin{table}[htbp]
\caption{Comparison of representative training-set construction strategies
for blood-vessel and gland segmentation. Values are Dice coefficients
(\%, mean $\pm$ SD) across five folds.}
\label{tab:representative_bv_gland}
\small
\begin{tabular*}{\textwidth}{@{\extracolsep{\fill}}llccc@{}}
\toprule
\textbf{Task} & \textbf{Strategy} & \textbf{Workload (\%)}
& \textbf{Validation Dice} & \textbf{Internal-Test Dice} \\
\midrule
BV
& Positive-only
& 16.4
& $47.80 \pm 4.80$
& $32.14 \pm 3.98$ \\
& Random-balanced (1:1)
& 32.8
& $57.84 \pm 4.06$
& $48.53 \pm 8.16$ \\
& All-tissue
& 100.0
& \textbf{$62.46 \pm 6.63$}
& \textbf{$60.38 \pm 9.76$} \\
& FTU-Seek
& 50.0
& $60.11 \pm 7.43$
& $57.68 \pm 6.88$ \\
\midrule
Gland
& Positive-only
& 36.9
& $80.78 \pm 11.30$
& $80.01 \pm 2.46$ \\
& Random-balanced (1:1)
& 73.78
& $87.93 \pm 4.51$
& $85.39 \pm 0.57$ \\
& All-tissue
& 100.0
& \textbf{$89.24 \pm 1.61$}
& \textbf{$86.57 \pm 0.69$} \\
& FTU-Seek
& 67.2
& $88.48 \pm 2.89$
& $86.33 \pm 0.27$ \\
\bottomrule
\end{tabular*}

\vspace{1ex}
\noindent{\footnotesize{Random-balanced training used an equal number of randomly selected
negative patches and positive patches. FTU-Seek Top1000 was used as a representative operating point for
cross-task comparison. The~same candidate Top$K$ values were evaluated across
the FTU categories to characterize the accuracy--workload trade-off under
matched sampling budgets; these values were exploratory operating points and
were not selected using a prevalence-adaptive rule or intended to define a
universally optimal budget. Neither the BV nor the gland cohort included a held-out cohort;
therefore, only development-set validation and internal-test results are
reported. Workload denotes the proportion of retained training patches
relative to all-tissue training.}}
\end{table}

\begin{table}[htbp]
\caption{Segmentation performance across hard-negative budgets for blood-vessel and gland segmentation. FTU-Seek values are reported as mean $\pm$ SD across five folds, whereas Random values are summarized as mean $\pm$ SD across five random seeds, with~each seed evaluated using \mbox{five-fold~models.}}
\label{tab:budget_bv_gland}
\small
\begin{tabular*}{\textwidth}{@{\extracolsep{\fill}}lccccc@{}}
\toprule
\textbf{Task} & \boldmath{$K$}
& \begin{tabular}[c]{@{}c@{}}\textbf{FTU-Seek}\\ \textbf{Validation}\end{tabular}
& \begin{tabular}[c]{@{}c@{}}\textbf{Random Validation}\\ \textbf{(5 Seeds)}\end{tabular}
& \begin{tabular}[c]{@{}c@{}}\textbf{FTU-Seek}\\ \textbf{Internal}\end{tabular}
& \begin{tabular}[c]{@{}c@{}}\textbf{Random Internal}\\ \textbf{(5 Seeds)}\end{tabular} \\
\midrule
BV
& 100
& $51.50 \pm 6.92$
& $53.25 \pm 4.66$
& $41.14 \pm 10.32$
& $37.16 \pm 3.85$ \\
& 300
& $57.03 \pm 5.35$
& $58.63 \pm 3.29$
& $47.93 \pm 11.80$
& $47.23 \pm 6.78$ \\
& 500
& $58.41 \pm 6.76$
& $58.45 \pm 2.97$
& $50.59 \pm 8.41$
& $47.63 \pm 11.62$ \\
& 1000
& $60.11 \pm 7.43$
& $60.95 \pm 7.25$
& $57.68 \pm 6.88$
& $54.21 \pm 6.17$ \\
\midrule
Gland & 100
& $84.81 \pm 9.37$
& $87.32 \pm 4.12$
& $83.01 \pm 2.44$
& $84.43 \pm 0.53$ \\
& 300
& $87.78 \pm 3.38$
& $87.06 \pm 3.90$
& $85.13 \pm 0.38$
& $84.70 \pm 0.98$ \\
& 500
& $87.46 \pm 5.27$
& $87.77 \pm 4.19$
& $85.90 \pm 0.91$
& $85.47 \pm 1.03$ \\
& 1000
& $88.48 \pm 2.89$
& $88.59 \pm 2.36$
& $86.33 \pm 0.27$
& $86.49 \pm 0.82$ \\
\bottomrule
\end{tabular*}

\vspace{1ex}
\noindent{\footnotesize{Random denotes random selection of the same number of target-absent
patches per WSI as the corresponding FTU-Seek Top$K$ configuration.
Neither the BV nor the gland cohort included a held-out cohort; therefore,
only development-set validation and internal-test results are reported.}}
\end{table}

\begin{table}[htbp]
\caption{Comparison of representative TLS training-set construction~strategies.}
\label{tab:tls_representative_strategies}
\small
\begin{tabular*}{\textwidth}{@{\extracolsep{\fill}}lcccc@{}}
\toprule
\textbf{Strategy}
& \textbf{Workload (\%)}
& \begin{tabular}[c]{@{}c@{}}\textbf{Development-Set}\\ \textbf{Validation}\end{tabular} 
& \begin{tabular}[c]{@{}c@{}}\textbf{Internal}\\ \textbf{Test}\end{tabular}
& \begin{tabular}[c]{@{}c@{}}\textbf{Held-Out}\\ \textbf{TLS Cohort}\end{tabular} \\
\midrule
Positive-only
& 4.4
& $57.20 \pm 25.23$
& $73.55 \pm 4.61$
& $61.90 \pm 20.41$ \\
Random-balanced (1:1)
& 8.8
& $71.50 \pm 12.55$
& $82.25 \pm 2.11$
& $70.83 \pm 16.23$ \\
All-tissue
& 100.0
& $71.63 \pm 17.79$
& $84.85 \pm 1.60$
& $75.12 \pm 12.67$ \\
FTU-Seek
& 27.6
& $76.91 \pm 10.14$
& $85.12 \pm 2.08$
& $76.69 \pm 11.89$ \\
\bottomrule
\end{tabular*}

\vspace{1ex}
\noindent{\footnotesize{Random-balanced training used all positive patches and an equal number of negative patches randomly sampled from the target-absent patch pool, thereby maintaining a 1:1 positive-to-negative patch ratio. For the representative-strategy comparison, FTU-Seek used Top1000 as a conservative operating point for subsequent downstream analyses. The~different $K$ values were explored to characterize the accuracy--workload trade-off, rather than to define a universally optimal budget. The~internal test and held-out TLS cohort were not used for configuration selection. Workload denotes the proportion of retained training patches relative to all-tissue training.}}
\end{table}

For TLS segmentation, positive-only training used 4.4\% of the all-tissue
workload but achieved only $73.55 \pm 4.61\%$ independent-test Dice.
Random-balanced training improved the test Dice to $82.25 \pm 2.11\%$ with
8.8\% workload, whereas all-tissue training reached
$84.85 \pm 1.60\%$. FTU-Seek Top$K$ sampling matched or exceeded the
all-tissue test performance at three of the four evaluated budgets while
using substantially fewer training patches. FTU-Seek Top100 achieved
$85.08 \pm 1.07\%$ Dice using 6.3\% of the patches, and~FTU-Seek Top500
achieved the highest independent-test Dice of $85.14 \pm 1.61\%$ using
15.8\% of the workload. Compared with matched random Top$K$ sampling,
FTU-Seek improved the internal-test Dice by 5.22, 3.81, 3.48, and~1.20 percentage points at Top100, Top300, Top500, and~Top1000,
respectively. The~corresponding paired slide-level comparisons on the
independent held-out TLS cohort are reported separately in
Table~\ref{tab:tls_heldout_paired}.

\begin{table}[htbp]
\caption{Paired slide-level comparisons between FTU-Seek and alternative
training-set construction strategies on the held-out TLS cohort.}
\label{tab:tls_heldout_paired}
\footnotesize
\begin{tabular*}{\textwidth}{@{\extracolsep{\fill}}lcccccc@{}}
\toprule
\textbf{Comparison}
& \begin{tabular}[c]{@{}c@{}}\textbf{FTU-Seek}\\ \textbf{Dice}\end{tabular}
& \begin{tabular}[c]{@{}c@{}}\textbf{Comparator}\\ \textbf{Dice}\end{tabular}
& \begin{tabular}[c]{@{}c@{}}\textbf{Mean}\\ \boldmath{$\Delta$}\end{tabular}
& \textbf{95\% CI}
& \begin{tabular}[c]{@{}c@{}}\textbf{Paired}\\ \boldmath{$t$}\end{tabular}
& \textbf{Wilcoxon} \\
\midrule
FTU-Seek Top100 vs. Random Top100
& $73.99 \pm 13.87$
& $71.41 \pm 15.43$
& $+2.58$
& $[1.14,\ 4.02]$
& $0.00096$
& $0.00021$ \\
FTU-Seek Top300 vs. Random Top300
& $76.54 \pm 13.03$
& $67.59 \pm 18.74$
& $+8.96$
& $[5.61,\ 12.31]$
& $<$0.0001
& $<$0.0001 \\
FTU-Seek Top500 vs. Random Top500
& $75.19 \pm 13.54$
& $70.48 \pm 16.84$
& $+4.72$
& $[3.06,\ 6.38]$
& $<$0.0001
& $<$0.0001 \\
FTU-Seek Top1000 vs. Random Top1000
& $76.69 \pm 11.89$
& $72.97 \pm 15.32$
& $+3.72$
& $[2.05,\ 5.38]$
& $<$0.0001
& $<$0.0001 \\
FTU-Seek Top1000 vs. All-tissue
& $76.69 \pm 11.89$
& $75.12 \pm 12.67$
& $+1.57$
& $[0.86,\ 2.28]$
& $<$0.0001
& $<$0.0001 \\
FTU-Seek Top1000 vs. Random-balanced (1:1)
& $76.69 \pm 11.89$
& $70.83 \pm 16.23$
& $+5.86$
& $[3.64,\ 8.07]$
& $<$0.0001
& $<$0.0001 \\
\bottomrule
\end{tabular*}

\vspace{1ex}
\noindent{\footnotesize{For the held-out TLS cohort, Table~\ref{tab:tls_all_evaluation_sets}
reports aggregate performance summaries across cross-validation folds and random seeds,
whereas the present table reports WSI-level paired analyses. Specifically, Dice was
calculated separately for each of the 30 held-out WSIs, with~each WSI serving as the
statistical pairing unit. Therefore, the~present analysis estimates within-slide
performance differences rather than reproducing the aggregate performance summaries
reported in Table~\ref{tab:tls_all_evaluation_sets}, and~the corresponding mean values
are not expected to be numerically identical.}}
\end{table}

BV segmentation showed a stronger dependence on negative-patch coverage. Positive-only training achieved only $32.14 \pm 3.98\%$ test Dice, whereas all-tissue training achieved the best performance ($60.38 \pm 9.76\%$). FTU-Seek performance increased as the hard-negative budget expanded, rising from $41.14 \pm 10.32\%$ at Top100 to $57.68 \pm 6.88\%$ at Top1000. The~latter used 50.0\% of the all-tissue workload and approached all-tissue performance while halving the retained training patches. FTU-Seek also outperformed matched random Top$K$ sampling at all BV budgets, with~test Dice gains of 3.98, 0.70, 2.96, and~3.47 percentage points from Top100 to~Top1000.

For gland segmentation, all strategies achieved relatively high Dice scores, consistent with the greater abundance and more distinct morphology of glandular structures. All-tissue training remained the strongest baseline ($86.57 \pm 0.69\%$ test Dice), but~reduced-workload strategies retained most of this performance. FTU-Seek Top300, Top500, and~Top1000 achieved $85.13 \pm 0.38\%$, $85.90 \pm 0.91\%$, and~$86.33 \pm 0.27\%$, respectively. 
FP-find Top1000 was within 0.24 percentage points of all-tissue training while reducing the workload to 67.2\%. However, FP-find did not consistently outperform matched random Top$K$ sampling across gland budgets, with~random sampling performing comparably or better at several settings. These results indicate that FP-find offers no clear overall advantage in gland segmentation, likely because gland-containing patches are relatively abundant and morphologically easier to~distinguish.

As summarized in \Cref{tab:tls_all_evaluation_sets,tab:representative_bv_gland,tab:budget_bv_gland}, the~value of hard-negative mining was task-dependent. TLS benefited most from classifier-guided selection at small workloads, BV required broader negative coverage to approach all-tissue performance, and~gland segmentation was comparatively insensitive to the negative-sampling strategy. Overall, FTU-Seek provided the clearest advantage for sparse or morphologically ambiguous FTUs, where randomly selected background patches are less likely to capture difficult false-positive~regions.

\subsection{Multi-Seed Stability of Random-Sampling~Baselines}
We next examined whether the performance of the random-sampling baselines depended strongly on a single random draw. Across TLS, blood-vessel, and~gland segmentation, each random-sampling configuration was repeated using four additional seeds, resulting in five seeds in total when the original seed was included. The~repeated runs showed limited-to-moderate variability across random seeds, with~the magnitude depending on the FTU category and negative-patch budget.
For TLS segmentation, the~random Top100, Top300, Top500, and~Top1000 baselines achieved mean Dice values of $79.86 \pm 3.45\%$, $80.88 \pm 1.55\%$, $81.66 \pm 1.24\%$, and~$83.92 \pm 0.54\%$, respectively, on~the internal two-WSI test cohort. The~corresponding FP-find results from the original seed-1 experiments were 85.08\%, 84.69\%, 85.14\%, and~85.12\%. Although~these comparisons should be interpreted cautiously because the internal test cohort contained only two WSIs, the~FP-find results remained above the random-baseline averages at all four~budgets.
 
The benefit was less pronounced for the denser FTU tasks. For~blood-vessel segmentation, the~differences between FP-find and the random-baseline averages were $+3.98$, $+0.70$, $+2.96$, and~$+3.47$ percentage points from Top100 to Top1000, respectively. For~gland segmentation, the~corresponding differences were $-1.42$, $+0.43$, $+0.43$, and~$-0.16$ percentage points. These results are consistent with the task-dependent behavior observed in the main experiments: random negative sampling may be sufficient when target-containing patches are relatively abundant and morphologically distinct, whereas classifier-guided selection is more advantageous for sparse or morphologically ambiguous~targets.

\subsection{Evaluation on an Independent Held-Out TLS~Cohort}

To determine whether the observed performance differences were preserved beyond the original two-slide internal test set, we evaluated all 11 segmentation strategies on an independent held-out cohort of 30 TLS WSIs from the same institution. The~cohort included 30 patients and 30 WSIs, with~1184 annotated TLS instances, 123,484 tissue patches, and~a TLS pixel-area ratio of 1.68\% (\Cref{tab:tls_heldout_characteristics}). This cohort was not used during training, model selection, Top$K$ selection, or~threshold tuning. The~evaluated strategies included positive-only training, all-tissue training, random-negative sampling, random Top$K$ sampling, and~classifier-guided FP-find Top$K$ sampling.

\begin{table}[htbp]
\caption{Characteristics and spatial sparsity of the independent held-out TLS~cohort.}
\label{tab:tls_heldout_characteristics}
\begin{tabular}{lr}
\toprule
\textbf{Characteristic} & \textbf{Value} \\
\midrule
Patients & 30 \\
WSIs & 30 \\
Total TLS instances & 1184 \\
Mean TLS instances per WSI & 39.47 \\
Median TLS instances per WSI & 33.00 \\
Total tissue patches & 123,484 \\
TLS-containing patches & 9231 \\
TLS-absent patches & 114,253 \\
TLS-containing patch ratio (\%) & 7.48 \\
Tissue pixel area & 525,755,637 \\
TLS pixel area & 8,822,896 \\
TLS pixel area ratio (\%) & 1.68 \\
\bottomrule
\end{tabular}
\noindent{\footnotesize{The cohort contained one WSI per patient. Tissue patches were extracted at
1~$\unit{\micro\metre}$/pixel with a nominal patch size of $256 \times 256$
pixels before resizing to the model input resolution. TLS-containing patches
overlapped at least one annotated TLS region, whereas TLS-absent patches did
not overlap an annotated TLS region. Pixel areas were calculated from the
corresponding tissue and TLS annotation masks.}}
\end{table}

On the held-out TLS cohort, FTU-Seek achieved slide-level Dice values of
73.99\%, 76.54\%, 75.19\%, and~76.69\% at Top100, Top300, Top500, and~Top1000, respectively. The~corresponding matched random Top$K$ strategies
achieved 71.41\%, 67.59\%, 70.48\%, and~72.97\%, respectively. The~largest
paired improvement was observed at Top300, whereas Top1000 achieved the
highest absolute FTU-Seek~Dice.

Based on the development-set analysis, Top1000 was retained as a
pre-specified representative operating point for comparisons with all-tissue
and random-balanced training. The~held-out cohort was not used for selecting
the operating point or tuning the segmentation procedure. Formal paired
comparisons across the 30 held-out WSIs are reported in Table~\ref{tab:tls_heldout_paired}.

\subsection{Paired Slide-Level Analysis on the Held-Out TLS~Cohort}

To assess whether the observed benefit of FTU-Seek was preserved in an independent held-out cohort,
we performed paired slide-level analyses on the held-out TLS cohort. The~same
30 WSIs were evaluated using FTU-Seek and the corresponding comparator
strategy. For~each WSI, the~paired Dice difference was calculated as
\[
\Delta_i =
\mathrm{Dice}_{\mathrm{FTU\text{-}Seek},i}
-
\mathrm{Dice}_{\mathrm{Comparator},i},
\]
where a positive value indicates better performance of FTU-Seek. Each model
prediction was generated using a five-fold probability ensemble, in~which the
foreground probabilities from the five fold models were averaged before
thresholding. The~mean paired difference, its two-sided 95\% confidence
interval, paired $t$-test, and~Wilcoxon signed-rank test were calculated
across the 30 WSIs. The~internal test cohort contained only two WSIs and was
therefore excluded from formal slide-level hypothesis testing and retained
for descriptive comparison only. The~paired comparisons included matched
random Top$K$ sampling at all four budgets, as~well as the representative
Top1000 comparisons with all-tissue and random-balanced~training.

Across all evaluated hard-negative budgets, FTU-Seek achieved higher
slide-level Dice than matched random Top$K$ sampling on the held-out TLS
cohort. The~mean paired improvements ranged from 2.58 to 8.96 percentage
points, and~all corresponding 95\% confidence intervals excluded zero. At~the
representative Top1000 operating point, FTU-Seek also showed positive paired
differences relative to all-tissue training ($\Delta=+1.57$ percentage points)
and random-balanced training ($\Delta=+5.86$ percentage points). These results
provide independent support for the generalization of FTU-Seek beyond the
development data. The~largest improvement over matched random sampling was
observed at Top300, whereas Top1000 achieved the highest absolute FTU-Seek
Dice.

\subsection{Qualitative~Analysis}

\begin{figure}[H]
    \centering
    \includegraphics[width=1.0\textwidth]{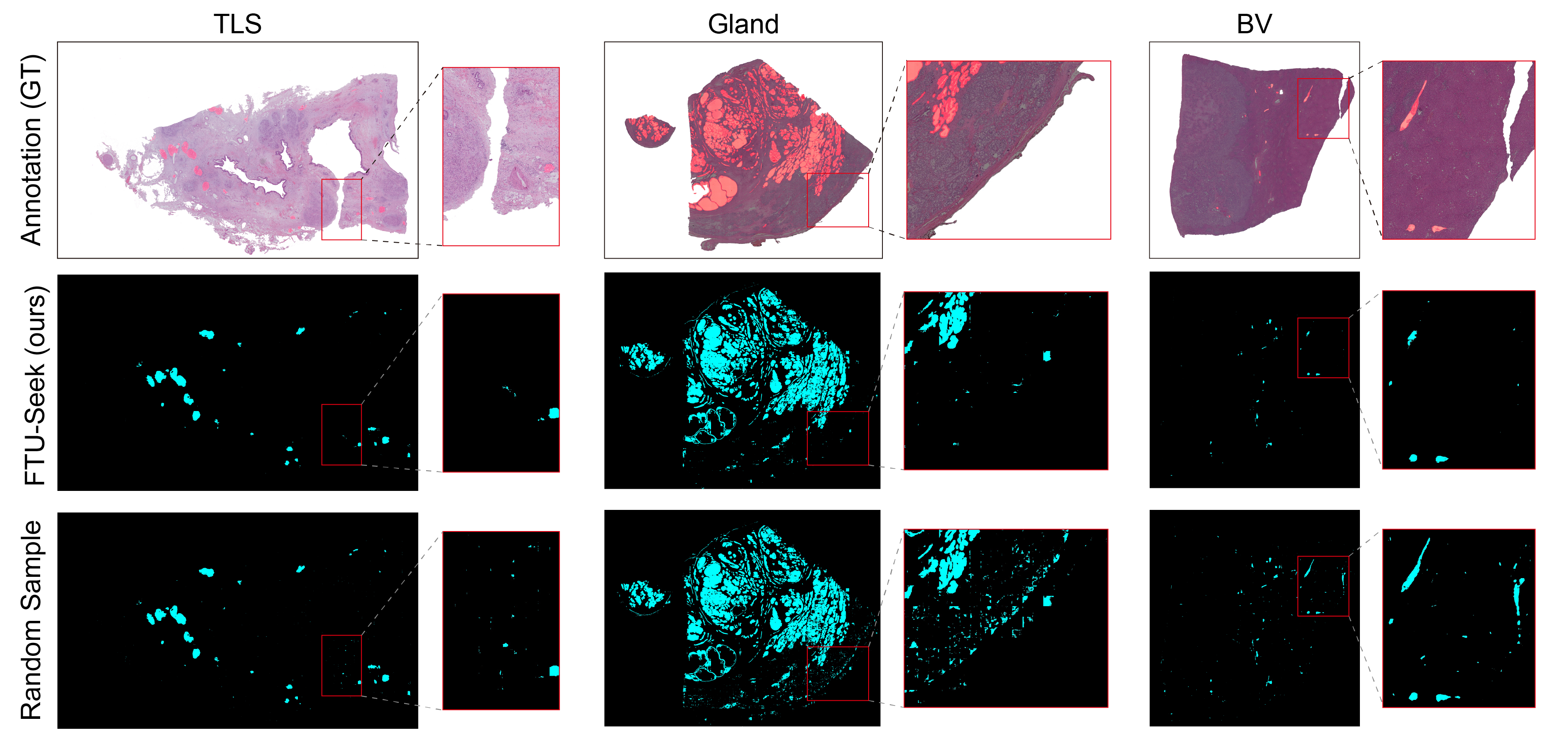}
    \caption{Visual comparison of segmentation performance across the three FTU cohorts. Representative results are shown for TLS, gland, and~blood vessel segmentation using classifier-guided FTU-Seek and matched random negative sampling. Red boxes indicate the magnified regions used to highlight false-positive predictions and boundary discrepancies. GT, ground~truth.}
    \label{fig:results_vision}
\end{figure}

\Cref{fig:results_vision} presents representative whole-slide segmentation results for the TLS, gland, and~BV cohorts, together with magnified regions for detailed comparison. Across the three tasks, FTU-Seek produced segmentation patterns that were generally more consistent with the ground-truth annotations than those obtained using random negative sampling. The~differences were most evident in regions containing sparse targets, morphologically ambiguous background structures, or~complex~boundaries.

For TLS segmentation, both methods identified the major TLS regions distributed across the tissue section. However, random negative sampling generated additional scattered predictions in background regions, particularly within the enlarged region, whereas FTU-Seek more effectively suppressed these isolated false-positive detections. At~the same time, the~principal TLS foci were retained, indicating that classifier-guided hard-negative selection improved background discrimination without substantially reducing sensitivity to sparse target~regions.

For gland segmentation, the~overall spatial distribution of glandular tissue was captured by both approaches. Nevertheless, random negative sampling produced a visibly denser and more fragmented prediction pattern in the magnified region, including responses extending into non-glandular tissue. In~comparison, FTU-Seek yielded a cleaner segmentation map with fewer spurious structures and a distribution more closely aligned with the annotated gland regions. These observations are consistent with its ability to prioritize morphologically confusing negative patches during~training.

The distinction between the two strategies was also apparent in BV segmentation. Although~both methods detected major vascular structures, random negative sampling produced several elongated or punctate false-positive regions in the background. FTU-Seek suppressed many of these vessel-like confounders while preserving the principal annotated vessels. This behavior is particularly important for BV segmentation, where tissue clefts, elongated stromal structures, and~other linear components may resemble vascular~profiles.

Overall, the~qualitative results support the quantitative findings in \Cref{tab:tls_all_evaluation_sets,tab:representative_bv_gland,tab:budget_bv_gland} and \Cref{fig:results_vision}. Compared with random negative sampling, FTU-Seek primarily improved segmentation by reducing false-positive predictions and refining target boundaries, rather than by simply increasing the total predicted area. The~benefit was especially evident for sparse or morphologically ambiguous FTUs, demonstrating that classifier-guided hard-negative sampling provides more informative background examples for segmentation model~training.

\subsection{Proof-of-Concept Downstream Phenotypic Analyses Enabled by FTU-Seek~Segmentation}
To illustrate the utility of FTU-Seek beyond segmentation performance, we converted segmentation masks from external TCGA cohorts into quantitative FTU phenotypes and examined their relationships with selected biological and clinicopathological characteristics. These proof-of-concept analyses were exploratory and were intended to demonstrate that FTU-Seek outputs can support downstream phenotyping and hypothesis generation, rather than to develop or validate clinical prediction~models.

\begin{figure}[H]
    \centering
    \includegraphics[width=.99\textwidth]{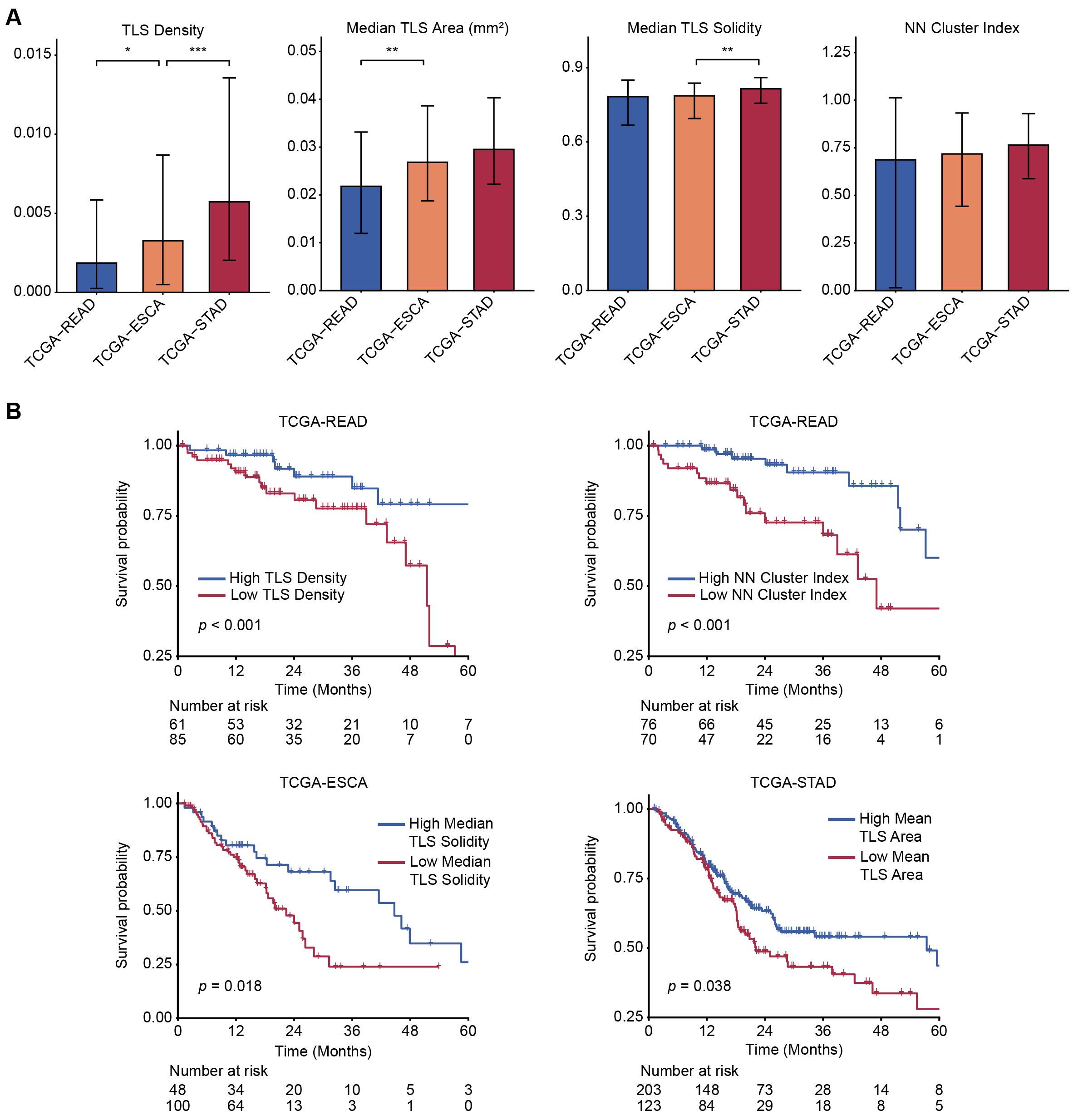}
    \caption{Exploratory characterization of TLS phenotypes and survival associations in TCGA cohorts. 
   (\textbf{A}) Comparison of multi-dimensional TLS features across TCGA-READ, TCGA-ESCA, and~TCGA-STAD. Panels from left to right present TLS Density, Median TLS area, median TLS solidity, and~NN (nearest neighbor) cluster index. * \textit{p} < 0.05, ** \textit{p} < 0.01, and *** \textit{p} < 0.001.
   (\textbf{B}) Kaplan--Meier survival analysis of key TLS features. Patients were stratified into high- and low-value groups based on exploratory data-derived cutoff values. Represented are TLS density and NN cluster index in the TCGA-READ cohort, median TLS solidity in the TCGA-ESCA cohort, and~Mean TLS area in the TCGA-STAD cohort.}
\label{fig:tls_vision}
\end{figure}

For TLSs, segmentation-derived phenotypes were quantified in TCGA-READ, TCGA-ESCA, and~TCGA-STAD. Matched survival analyses included 146 READ, 148 ESCA, and~326 STAD patients after merging available WSI-derived TLS features with clinical outcomes. A~total of 60 TLS features captured complementary aspects of immune organization, including TLS burden, size composition, hotspot density, and~nearest-neighbor clustering. Across cohorts, mean TLS density ranged from 0.0048 in READ to 0.0116 in STAD, indicating substantial inter-cohort heterogeneity in TLS abundance (\Cref{fig:tls_vision}A). Furthermore, TLS features exhibited significant heterogeneity across the different cohorts (\Cref{fig:tls_vision}A). Specifically, the~median instance-level TLS area in the READ cohort was significantly smaller than that in the other cohorts ($p$ = 0.010). Conversely, TLSs in the STAD cohort demonstrated higher solidity ($p$ = 0.001). In~morphological analysis, solidity is defined as the ratio of the actual TLS area to the area of its convex hull, essentially reflecting a more compact structure with smoother, less irregular boundaries. Regarding spatial distribution, there were no significant differences across the three cohorts in the nearest neighbor (NN) cluster index. This spatial metric quantifies the degree of TLS clustering by comparing the observed average distance between adjacent TLSs against the expected distance under a completely random spatial distribution. Specifically, in~READ, a~1-SD increase in TLS density was associated with a 51\% reduction in the risk of death (\mbox{HR = 0.49}, 95\% CI: 0.26--0.92; nominal $p = 0.026$; FDR-adjusted $p = 0.25$), whereas higher NN cluster index values were associated with a 60\% lower risk (HR = 0.40, 95\% CI: 0.17--0.95; nominal $p = 0.039$; FDR-adjusted $p = 0.25$). Neither association met the exploratory FDR threshold. 
For illustrative exploratory visualization, Kaplan--Meier curves were additionally generated for selected TLS phenotypes using data-derived cut points. These included TLS density and NN cluster index in READ, median TLS solidity in ESCA, and~mean TLS area in STAD (\Cref{fig:tls_vision}B). The~resulting group separations were used solely to illustrate potential survival stratification and were not interpreted as validated or multiplicity-adjusted prognostic~associations.

\begin{figure}[H]
    \centering
    \includegraphics[width=1.0\textwidth]{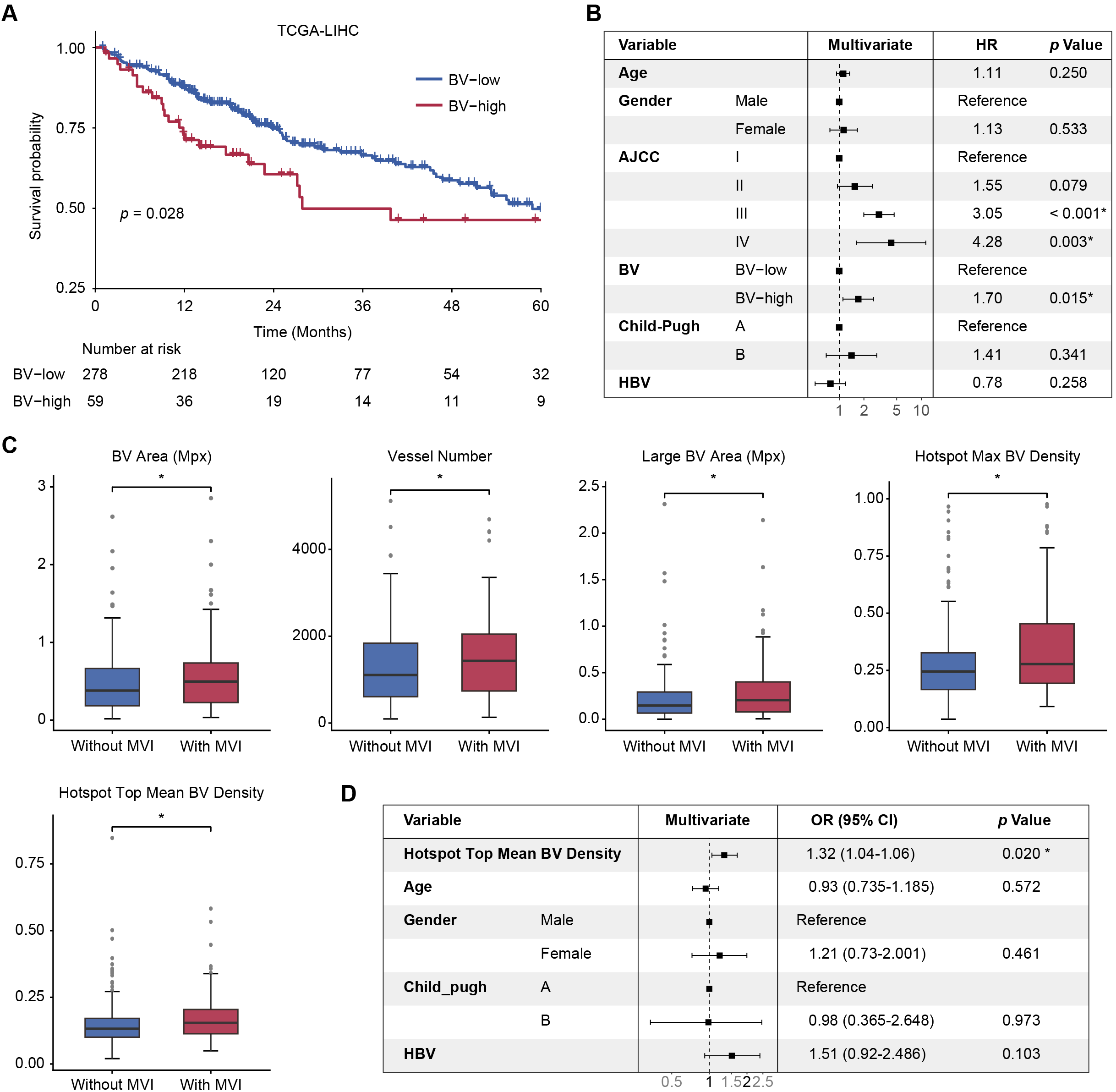}
    \caption{Exploratory
 associations of blood-vessel phenotypes with overall survival (OS) and microvascular invasion (MVI) in TCGA-LIHC. 
   (\textbf{A}) Kaplan--Meier estimates of overall survival (OS) for patients stratified into BV-low (blue line) and BV-high (red line) groups based on an exploratory data-derived cutoff value. Statistical significance was assessed using the log-rank test.
(\textbf{B}) Forest plot of the multivariable Cox proportional hazards regression analysis for OS. The~model incorporates clinical covariates including age, gender, AJCC stage, BV classification (BV-low vs. BV-high), Child--Pugh score, and~Hepatitis B Virus (HBV) status. Hazard ratios (HR) and corresponding $p$-values are presented.
(\textbf{C}) Boxplots displaying the distribution of quantitative computational BV features in patient groups without and with MVI. The~central horizontal line represents the median, box edges denote the interquartile range (IQR).
(\textbf{D}) Forest plot presenting the multivariable logistic regression analysis for MVI status. The~model evaluates hotspot top mean BV density and adjusts for baseline clinical characteristics (age, gender, Child--Pugh score, and~HBV status). Odds ratios (ORs), 95\% confidence intervals (CIs), and~$p$-values are provided for each variable. $^* p < 0.05$.} 
\label{fig:bv}
\end{figure}

For blood vessels, FTU-Seek-derived vascular phenotypes were quantified in TCGA-LIHC. The~LIHC cohort included 337 patients and 121 death events. Among~the 36~vascular phenotypes evaluated, higher BV count density showed a potential adverse association with OS (log-rank $p$ = 0.028; \Cref{fig:bv}A). In~the multivariable Cox analysis, the~high-BV-density group showed a higher mortality risk than the low-density group after adjustment for age, clinical stage, Child--Pugh class, and~HBV status (adjusted HR = 1.70, 95\% CI: 1.10--2.59; $p$ = 0.016; \Cref{fig:bv}B). The~proportional-hazards diagnostic indicated evidence of a time-varying association, and~the HR was therefore interpreted as an exploratory overall summary. No substantial multicollinearity was observed among the included predictors (all VIFs $<$ 1.15). Furthermore, we observed significant differences in the distribution of computationally extracted vascular features---including BV area (nominal $p = 0.032$; FDR-adjusted $p = 0.087$), large BV area (nominal $p = 0.025$; FDR-adjusted $p = 0.087$), hotspot top mean BV density (nominal $p = 0.017$; FDR-adjusted $p = 0.087$), and~hotspot max BV density (nominal $p = 0.011$; FDR-adjusted $p = 0.087$)---between patients with and without MVI (\Cref{fig:bv}C). Notably, in~the multivariable logistic regression analysis, each 1-SD increase in hotspot top mean BV density---the mean BV density within the top decile of vascular hotspots---was associated with higher odds of MVI (adjusted OR = 1.32, 95\% CI: 1.04--1.66; $p$ = 0.020; \Cref{fig:bv}D). RCS analysis further supported an overall association between hotspot top mean BV density and MVI ($p_{\mathrm{overall}}$ = 0.001), with~evidence of nonlinearity ($p_{\mathrm{nonlinearity}}$ = 0.015) and an overall trend toward higher odds of MVI at greater BV density. Overall, these analyses illustrate how FTU-Seek-derived vascular phenotypes can be used to explore potential relationships between tumor-associated vascular organization, mortality risk, and~MVI.

\begin{figure}[H]
    \centering
    \includegraphics[width=1.0\textwidth]{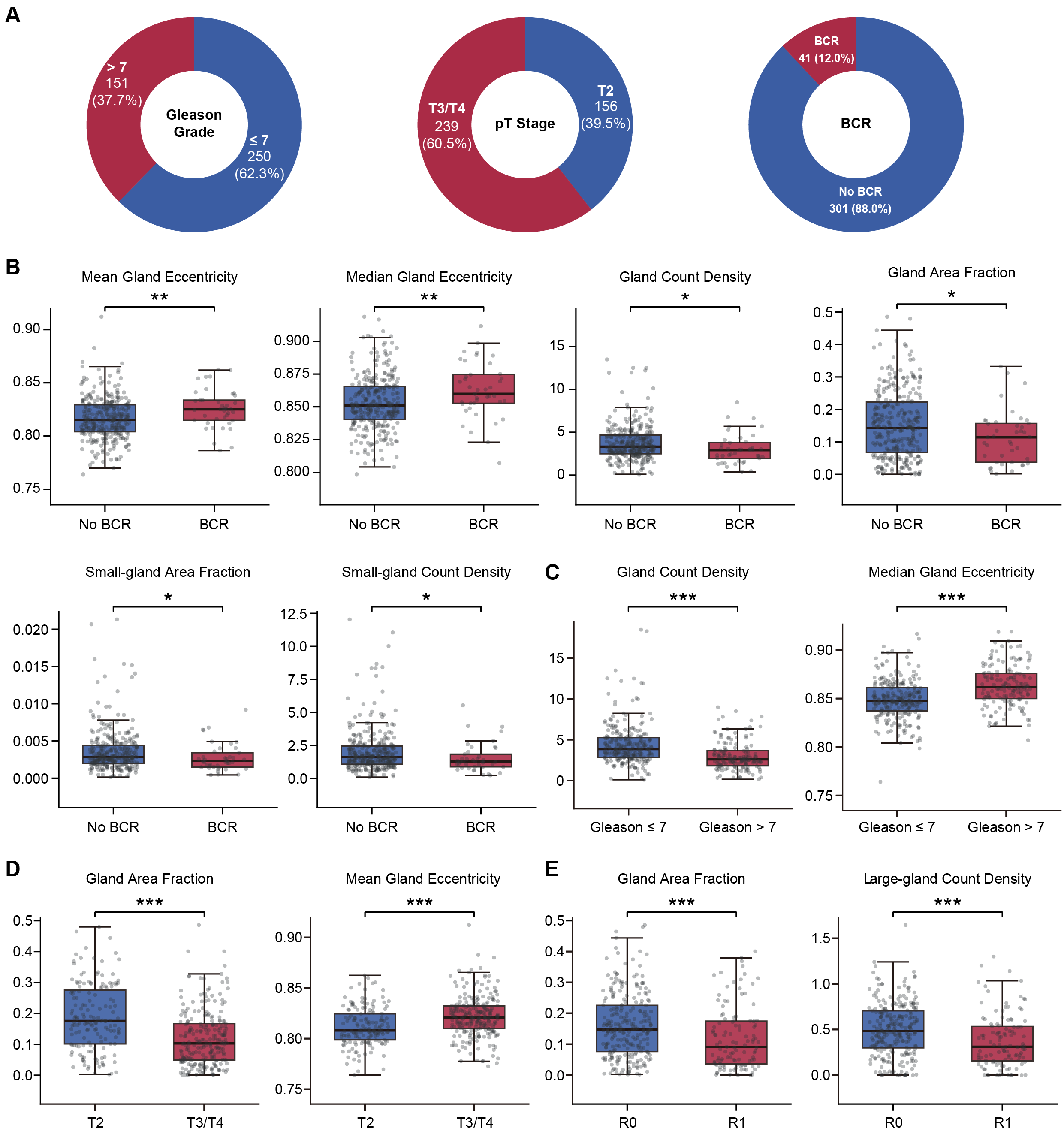}
    \caption{Association of segmented gland features with key clinicopathological determinants in the TCGA-PRAD cohort.
(\textbf{A}) Donut charts illustrating the proportion of patients within the TCGA-PRAD cohort stratified by key clinical and pathological categories: Gleason grade ($\le$7 vs. $>$7), pT stage (T2~vs.~T3/T4), and~biochemical recurrence (BCR) status (No BCR vs. BCR). Slices indicate the count and percentage for each respective subgroup.
(\textbf{B}--\textbf{E}) Boxplots detailing the distribution of extracted gland features that exhibited statistically significant differences between the respective clinical groups.
(\textbf{B}) Distribution of gland features (mean gland eccentricity, median gland eccentricity, gland count density, gland area fraction, small-gland area fraction, and~small-gland count density) compared between patients without and with BCR.
(\textbf{C}) Comparison of gland count density and median gland eccentricity between Gleason grade $\le$ 7 and $>$7 groups.
(\textbf{D}) Distribution of gland area fraction and mean gland eccentricity across pT stage categories (T2 vs. T3/T4).
(\textbf{E}) Comparison of gland area fraction and large-gland count density categorized by resection margin status (R0 vs. R1).
For all boxplots, the~central horizontal line represents the median, the~box edges span the interquartile range (IQR).  $^*$ nominal $p < 0.05$, $^{**}$ nominal $p < 0.01$, $^{***}$ nominal $p < 0.001$.}
\label{fig:gland}
\end{figure}

FTU-Seek-derived 40 glandular architectural phenotypes were evaluated in TCGA-PRAD. The~baseline clinical and pathological characteristics of the study cohort are summarized in \Cref{fig:gland}A. In~terms of histological grading, the~majority of patients presented with a Gleason score of $\leq$ 7 (n = 250, 62.3\%), while 151 patients (37.7\%) had a high-grade disease with a Gleason score of > 7. Regarding pathological staging, a~higher proportion of patients were diagnosed with advanced pT stage (T3/T4, n = 239, 60.5\%) compared to localized stage (T2, n = 156, 39.5\%). Furthermore, at~the end of the follow-up period, BCR was observed in 41 patients (12.0\%), with~the remaining 301 patients (88.0\%) staying BCR-free (\Cref{fig:gland}A). FTU-Seek-derived glandular architectural phenotypes were evaluated in TCGA-PRAD. In~the BCR analysis, recurrent cases showed higher eccentricity and lower gland count density (nominal $p < 0.05$; FDR-adjusted $p < 0.1$; \Cref{fig:gland}B). Consistent with these findings, compared with tumors with a Gleason score $\leq$ 7, tumors with a Gleason score $>$ 7 had lower gland count density (median 2.60 vs. 3.86 glands/mm²; FDR-adjusted $p < 0.001$) and higher median gland eccentricity (median 0.862 vs. 0.847; FDR-adjusted $p < 0.001$; \Cref{fig:gland}C). Similar trends were observed for local extension: pT3/T4 tumors showed lower gland area (median 0.103 vs. 0.175; FDR-adjusted $p < 0.001$) and higher gland eccentricity (median 0.821 vs. 0.808; FDR-adjusted $p < 0.001$) than pT2 tumors \mbox{(\Cref{fig:gland}D)}. Additionally, positive surgical margins were associated with lower large-gland count density and lower total gland density (both FDR-adjusted $p < 0.001$; \Cref{fig:gland}E). Collectively, these findings demonstrate that FTU-Seek-derived glandular phenotypes capture architectural alterations associated with adverse pathological characteristics and disease recurrence, supporting their utility for quantitative downstream tissue~analysis.
\vspace{-6pt}

\section{Discussion}

In this study, we developed FTU-Seek
, a~pathology foundation model (PFM)-guided hard-negative learning framework for sparse functional tissue unit (FTU) segmentation in whole-slide images (WSIs). Unlike conventional approaches that primarily employ foundation models as feature extractors or segmentation backbones, FTU-Seek additionally exploits pretrained histomorphological representations to construct morphology-aware segmentation training sets~\cite{chen2024towards,liang2026separation,vorontsov2024foundation}. A~patch-level classifier first ranks FTU-absent tissue according to its predicted target-containing probability, after~which a static Top$K$ strategy selects the most target-like negative patches for segmentation training. This design addresses a key challenge in quantitative histopathology: biologically meaningful microscopic structures are often sparsely distributed throughout large tissue sections, whereas the majority of tissue patches contain redundant background and only a limited subset comprises morphologically confusing regions likely to generate false-positive predictions. Across TLS, gland, and~blood-vessel segmentation, our results show that morphology-aware negative selection improves the trade-off between segmentation accuracy and retained training workload, with~the clearest benefits observed for sparse or morphologically ambiguous FTUs. The~resulting segmentation outputs can further be transformed into quantitative tissue phenotypes for exploratory downstream~analysis.

The segmentation experiments highlight that the composition of the negative training set is more important than its absolute size. Training using only FTU-containing patches consistently resulted in inferior performance, confirming that representative negative tissue is indispensable for learning robust decision boundaries. Conversely, all-tissue training provides abundant negative information but requires processing a very large number of easy background patches that contribute little additional supervision. FTU-Seek addresses this imbalance by preferentially selecting classifier-ranked target-like negatives, thereby concentrating training on the background regions most likely to generate false-positive predictions. Although~this concept shares the objective of hard-example mining, FTU-Seek differs fundamentally from conventional online hard-example mining because difficult negatives are identified before segmentation training using frozen pathology foundation model representations rather than repeatedly re-mined during optimization. Consequently, the~proposed strategy is computationally reproducible, avoids repeated screening of the entire WSI patch pool, and~enables direct comparison with alternative sampling strategies under identical negative-sampling budgets.
The reduction in training workload should be distinguished from a reduction in peak GPU-memory usage. FTU-Seek decreases the total number of retained training patches and therefore improves training-data efficiency without requiring additional GPU memory or a larger computational platform. However, when the model architecture, batch size, input resolution, and~hardware configuration are fixed, reducing the total number of training patches does not necessarily reduce the peak GPU-memory usage per iteration. Accordingly, the~computational benefit of FTU-Seek should be interpreted as improved training-data efficiency under a fixed memory budget rather than as a proportional reduction in GPU-memory consumption. A~complete assessment of computational efficiency would additionally require systematic measurements of wall-clock training time, GPU utilization, peak memory, and~energy~consumption.

The task-dependent benefit of classifier-guided hard-negative selection may be explained by differences in target sparsity, class imbalance, and~morphological ambiguity. When targets are sparse and imbalance is severe, the~negative pool is dominated by easy background, making random sampling less likely to capture the small subset of target-like negatives most responsible for false-positive predictions. TLSs represented the sparsest segmentation task and exhibited the greatest efficiency gains. A~relatively small number of classifier-selected negatives was sufficient to recover or slightly exceed the performance of all-tissue training, suggesting that only a limited subset of inflammatory or stromal regions contributes substantially to false-positive TLS predictions. Blood-vessel segmentation demonstrated a different behaviour. Because~vessel-like structures occur in diverse morphological contexts, including tissue clefts, empty lumina, collagen bundles, elongated stromal regions, and~red blood cell-rich spaces, segmentation performance continued to improve as larger hard-negative pools were incorporated. Gland segmentation was comparatively insensitive to the sampling strategy because gland-containing patches were considerably more abundant, resulting in less severe class imbalance, and~the distinction between positive and negative tissue was substantially clearer. Collectively, these findings indicate that sparse FTUs derive greater benefit from targeted hard-negative selection, whereas more abundant FTUs with clearer class separation show smaller gains; broader morphological ambiguity may additionally require greater negative coverage. This task-dependent sampling strategy also provides a practical means of reducing redundant training data. By~removing easy-negative patches before segmentation training, FTU-Seek reduces the number of samples processed during optimization while keeping the segmentation architecture, batch size, input resolution, and~hardware configuration~unchanged.

The classifier probability distributions provide further insight into the mechanism underlying FTU-Seek. For~TLSs, the~vast majority of negative patches received probabilities close to zero, while only a small high-probability tail corresponded to localized target-like tissue. Such distributions naturally favour strict Top$K$ selection because informative negatives are concentrated within a relatively small subset. Blood vessels exhibited substantially broader overlap between positive and negative probability distributions, reflecting widespread morphological ambiguity and explaining why larger negative budgets were beneficial. Gland classification showed the clearest overall separation, although~intermediate-probability regions remained around gland boundaries, pseudo-glandular spaces, epithelial debris, and~fibrotic cavities. These observations indicate that classifier probabilities capture biologically meaningful morphological ambiguity rather than merely reflecting classification confidence. Future developments may therefore combine probability-based ranking with diversity constraints, uncertainty estimation, or~tissue-context clustering to obtain even more representative hard-negative~sets.

The patch-level classifier is an important intermediate component, but~its accuracy does not guarantee improved segmentation. Because~all target-containing patches are retained for segmentation training, classifier errors on positive patches do not directly remove positive training examples. In~contrast, errors within the target-absent pool alter the composition of the selected hard-negative set. If~a morphologically confusing negative patch receives an underestimated target-containing probability, it may be omitted from Top$K$ selection, leaving the corresponding false-positive pattern insufficiently represented during segmentation training. Conversely, overestimated probabilities may allocate Top$K$ slots to less informative negatives. Although~such selections may still increase the diversity of background examples, extensive misranking can reduce the efficiency of hard-negative sampling. Thus, the~quality and calibration of the patch-level classifier affect segmentation indirectly through negative-pool composition. Uncertainty-aware ranking or iterative negative re-mining may help mitigate this error propagation, but~would introduce additional computational cost and potential~variability.

An important component of FTU-Seek is the use of frozen pathology foundation model representations. Recent pathology foundation models such as UNI and Virchow have demonstrated remarkable transferability across computational pathology tasks by learning general histomorphological representations from large and heterogeneous pathology image collections~\cite{chen2024towards,liu2026computational,vorontsov2024foundation}. In~the present framework, these representations serve two complementary purposes. First, they provide robust patch-level features for identifying target-like negative tissue. Second, they supply transferable multi-scale representations for dense segmentation across structurally distinct FTUs. Freezing the encoder substantially reduces the number of trainable parameters while minimizing dependence on relatively small manually annotated datasets. Our results therefore suggest that pathology foundation models can contribute not only through transferable feature extraction but also through intelligent construction of segmentation training data, thereby extending their role beyond conventional transfer~learning.

Beyond computational efficiency, FTU-Seek enables quantitative characterization of tissue architecture. TLSs, blood vessels, and~glands represent complementary components of the tissue microenvironment that reflect adaptive immune organization, angiogenesis and vascular remodeling, and~epithelial differentiation, respectively. In~this study, their segmentation masks were converted into quantitative spatial phenotypes for exploratory downstream~analyses.

Our proof-of-concept downstream analyses further illustrate this potential. Quantitative TLS features spanning abundance, spatial organization, and~morphology exhibited variation across tumor types. The~observed differences among READ, ESCA, and~STAD are biologically plausible because TLS maturation, localization, cellular composition, and~interactions with the surrounding tumor microenvironment differ substantially across tumor types~\cite{cho2026pan,silicna2018germinal,yu2026chemotactic,xu2023heterogeneity,tang2025spatial}. Several TLS phenotypes also showed potential associations with OS when evaluated as continuous measures, supporting their utility for generating hypotheses regarding the relationship between immune architecture and patient outcome. Similarly, vascular phenotyping in hepatocellular carcinoma identified candidate features related to vessel density and localized vascular hotspots, with~exploratory associations observed for OS and MVI. Finally, in~prostate cancer, glandular phenotypes captured architectural alterations associated with adverse clinicopathological characteristics, including reduced gland density and increased eccentricity in more aggressive tumors. Collectively, these observations suggest that FTU-Seek-derived phenotypes retain information beyond the segmentation masks themselves and can serve as quantitative descriptors for subsequent biological and clinical~investigation.

Several limitations should be acknowledged. First, although~each development
cohort contained extensive pixel-level annotations, the~number of annotated
WSIs remained relatively limited, and~the internal test cohort contained only
two WSIs per segmentation task. Accordingly, the~internal-test results should
be interpreted as descriptive evidence rather than definitive estimates of
generalization. To~provide a stronger independent assessment, we additionally
evaluated the TLS segmentation strategies on 30 held-out WSIs from the same
institution. However, this cohort does not constitute multi-center external
validation. Larger multi-institutional studies will be necessary to evaluate
the robustness of FTU-Seek across diverse patient populations, staining
protocols, scanners, and~tissue-processing procedures~\cite{mongan2020checklist}.

The additional held-out analysis showed that FTU-Seek achieved higher
slide-level Dice than matched random Top$K$ sampling across all four evaluated
negative-patch budgets, with~positive paired differences and confidence
intervals excluding zero. These results support the robustness of the method
within the available same-institution data setting, while broader
multi-institutional validation remains~necessary.

Second, the~present workload metric was defined as the proportion of retained training patches relative to all-tissue training rather than direct measurements of computational time, GPU utilization, memory consumption, or~energy efficiency. Future studies should therefore include comprehensive computational benchmarking. Third, the~proposed static Top$K$ strategy primarily emphasizes false-positive suppression and does not explicitly incorporate false-negative regions, annotation uncertainty, or~difficult target boundaries. Iterative re-mining, uncertainty-guided sampling, boundary-aware optimization, and~combined false-positive/false-negative mining may further improve segmentation performance. 
Fourth, identical Top$K$ candidate values were evaluated across all FTU
categories, although~our results indicate that the optimal hard-negative
budget depends on target prevalence and morphological~complexity. 

The fixed per-WSI allocation also means that slides with different tissue
areas do not contribute negative patches in proportion to their available
tissue area. Proportional allocation and globally ranked negative-patch
selection were not evaluated in the present study. These alternative
allocation schemes, together with prevalence- or uncertainty-adaptive
budgets, remain directions for future~work.

Adaptive
sampling strategies based on classifier uncertainty, probability distributions,
or validation performance may therefore provide greater flexibility.
 Finally, FTU-Seek may have potential future applications in the analysis of stem cell-derived tissues, organoids, and~regenerative medicine specimens; however, these settings were not investigated in the present study and would require dedicated validation using appropriately annotated~datasets.

The downstream TCGA analyses should be interpreted as exploratory and hypothesis-generating. Because~manually annotated reference masks were unavailable, the~TCGA cohorts were used for exploratory downstream application rather than direct external validation of segmentation performance. These retrospective analyses were not designed for biomarker validation, and~the data-derived survival cut points should not be considered validated prognostic thresholds. Accordingly, the~findings primarily demonstrate the feasibility of deriving quantitative phenotypes from FTU-Seek segmentations. Future integration with spatial transcriptomics, proteomics, and~molecular pathology may further clarify the biological basis of these phenotypes and identify reproducible architectural features across disease~contexts.

Overall, FTU-Seek extends pathology foundation models beyond transferable feature representation by using pretrained histomorphological information to identify target-like negative tissue before segmentation training and construct compact, morphology-aware training sets. This pre-segmentation hard-negative selection reduces redundant background sampling and retained training workload while maintaining competitive segmentation performance, with~the clearest practical benefit observed for sparse and imbalanced FTUs. These findings should be interpreted in light of the relatively small number of annotated WSIs, the~lack of direct wall-clock and resource benchmarking, and~the dependence of the static Top$K$ strategy on classifier-based ranking and task-specific sampling budgets. Larger multi-institutional cohorts, systematic computational benchmarking, and~adaptive or iterative negative-selection strategies will be needed to establish the robustness and practical clinical applicability of~FTU-Seek.

\section{Conclusions}

Overall, FTU-Seek extends pathology foundation models beyond feature representation by using pretrained histomorphological information to identify target-like negative tissue before segmentation training and construct compact, morphology-aware training sets. This pre-segmentation hard-negative selection directly addresses the challenge of sparse FTU segmentation by reducing redundant background sampling and retained training workload while maintaining competitive segmentation performance, with~particularly clear benefits for sparse and imbalanced FTUs. The~present study is limited by the relatively small number of annotated WSIs and the absence of direct computational benchmarking, while the static TopK strategy remains dependent on classifier-based ranking and task-specific sampling requirements. Prospective validation in independent multi-institutional cohorts will therefore be required before the clinical applicability of FTU-Seek can be established, and~more adaptive sampling strategies warrant further~investigation.

\bibliography{1reference}

\end{document}